# CoNIC Challenge: Pushing the Frontiers of Nuclear Detection, Segmentation, Classification and Counting

Simon Graham[1,2,†,+], Quoc Dang Vu[1,†,+], Mostafa Jahanifar[1,+], Martin Weigert[3], Uwe Schmidt[4], Wenhua Zhang[5], Jun Zhang[6], Sen Yang[7], Jinxi Xiang[8], Xiyue Wang[9], Josef Lorenz Rumberger[10,11,12], Elias Baumann[13], Peter Hirsch[10,11], Lihao Liu[14], Chenyang Hong[15], Angelica I. Aviles-Rivero[14], Ayushi Jain[16,17], Heeyoung Ahn[18], Yiyu Hong[18], Hussam Azzuni[19], Min Xu[19], Mohammad Yaqub[19], Marie-Claire Blache[20], Benoît Piégu[20], Bertrand Vernay[21,22,23,24], Tim Scherr[25], Moritz Böhland[25], Katharina Löffler[25], Jiachen Li[26], Weiqin Ying[26], Chixin Wang[26], David Snead[2,27,28,+], Shan E Ahmed Raza[1,+], Fayyaz Minhas[1,+], Nasir M. Rajpoot[1,2,27,+] and the CoNIC Challenge Consortium[*]

[1]Tissue Image Analytics Centre, University of Warwick, Coventry, United Kingdom
[2]Histofy Ltd, Birmingham, United Kingdom
[3]Institute of Bioengineering, School of Life Sciences, EPFL, Lausanne, Switzerland
[4]Independent Researcher, Dresden, Germany
[5]The Department of Computer Science, The University of Hong Kong, Hong Kong
[6]Tencent AI Lab, Shenzhen, China
[7]College of Biomedical Engineering, Sichuan University, Chengdu, China
[8]Department of Precision Instruments, Tsinghua University, Beijing, China
[9]College of Computer Science, Sichuan University, Chengdu, China
[10]Max-Delbrueck-Center for Molecular Medicine in the Helmholtz Association, Berlin, Germany
[11]Humboldt University of Berlin, Faculty of Mathematics and Natural Sciences, Berlin, Germany
[12]Charité University Medicine, Berlin, Germany
[13]University of Bern, Bern, Switzerland
[14]Department of Applied Mathematics and Theoretical Physics, University of Cambridge, United Kingdom
[15]Department of Computer Science and Engineering, Chinese University of Hong Kong, Hong Kong
[16]Softsensor.ai, Bridgewater, New Jersey, United States of America
[17]PRR.ai, Texas, United States of America
[18]Department of R&D Center, Arontier Co. Ltd, Seoul, Republic of Korea
[19]Computer Vision Department, Mohamed Bin Zayed University of Artificial Intelligence, Abu Dhabi, United Arab Emirates
[20]CNRS, IFCE, INRAE, Université de Tours, PRC, 3780, Nouzilly, France
[21]Institut de Génétique et de Biologie Moléculaire et Cellulaire, Illkirch, France
[22]Centre National de la Recherche Scientifique, UMR7104, Illkirch, France
[23]Institut National de la Santé et de la Recherche Médicale, INSERM, U1258, Illkirch, France
[24]Université de Strasbourg, Strasbourg, France
[25]Institute for Automation and Applied Informatics Karlsruhe Institute of Technology Eggenstein-Leopoldshafen, Germany
[26]School of software engineering, South China University of Technology Guangzhou, China
[27]Department of Pathology, University Hospitals Coventry and Warwickshire NHS Trust, Coventry, United Kingdom
[28]Division of Biomedical Sciences, Warwick Medical School, University of Warwick, Coventry, United Kingdom

[†]First authors contributed equally
[+]Challenge organisers
[*]A list of consortium authors and their affiliations appear at the end of this paper

# Abstract

Nuclear detection, segmentation and morphometric profiling are essential in helping us further understand the relationship between histology and patient outcome. To drive innovation in this area, we setup a community-wide challenge using the largest available dataset of its kind to assess nuclear segmentation and cellular composition. Our challenge, named CoNIC, stimulated the development of reproducible algorithms for cellular recognition with real-time result inspection on public leaderboards. We conducted an extensive post-challenge analysis based on the top-performing models using 1,658 whole-slide images of colon tissue. With around 700 million detected nuclei per model, associated features were used for dysplasia grading and survival analysis, where we demonstrated that the challenge's improvement over the previous state-of-the-art led to significant boosts in downstream performance. Our findings also suggest that eosinophils and neutrophils play an important role in the tumour microevironment. We release challenge models and WSI-level results to foster the development of further methods for biomarker discovery.

# Introduction

Analysis of nuclei in a histopathology tissue slide can provide key information for identifying the presence or state of a disease. For example, their shape and appearance can be used to determine cancer grade, whereas the co-occurrence and distribution of different nuclei can be indicative of diagnosis and patient outcome. In particular, epithelial nuclear pleomorphism is a major component of the Nottingham Grading System for breast cancer[1], while increased amounts of immune cells may be a sign of certain conditions, such as inflammatory bowel disease[2,3]. Two particularly well-studied prognostic tissue-based biomarkers are Tumour-Infiltrating Lymphocytes (TILs)[4,5] and Cancer-Associated Fibroblasts (CAFs)[6,7]. Here, TILs have been linked to positive patient outcome and immunotherapy response, while CAFs are generally associated with poor outcome due to their role in promoting tumour development.

To assist with nuclear analysis, Deep Learning (DL) methods like HoVer-Net[9] and StarDist[10] have been used to automate nucleus recognition. However, these models require a large amount of labelled data to perform accurately. Obtaining pixel-level annotations is a time-consuming task that requires pathologist input, often leading to small datasets. To overcome this, recent semi-automatic methods involving pathologists have led to the collection of large datasets for nuclear segmentation. For example, PanNuke[11,12] collected point annotations of over 200,000 different nuclei with collaborating pathologists, and utilised a semi-automatic method for generating the nuclear boundaries[13]. The Lizard dataset[14] employed an iterative approach to label nearly half a million nuclei, using a combination of semi-automatic and manual pathologist-involved refinement steps to ensure accurate annotation. Datasets at such scales pave the way for the development and reliable evaluation of advanced DL models for nuclear recognition.

AI competitions have been pivotal in helping to drive forward the development of innovative DL models in Computational Pathology (CPath)[15-18], where carefully curated datasets are made available to participants around the world. However, even though there have been several previous competitions for automatic identification of nuclei in H&E images[19,20], all tend to suffer from a similar set of limitations. For example, the previous largest competition for nuclear segmentation and classification[21] used a dataset consisting of around 47 thousand nuclei, where only 15 thousand of these were used for evaluation. Furthermore, the evaluation images were available to participants, meaning that models could be tuned until a satisfactory visual performance was observed. Of course, this is not reflective of clinical practice and may ultimately lead to overfitting. Instead, it is desirable for images to be hidden from participants during evaluation to ensure reliable assessment of model performance and to minimise the risk of test data hacking.

Despite competitions being a great way of accelerating research for AI-based nuclear recognition, the ultimate aim is to enable the extraction of interpretable biomarkers and use them in downstream clinical tasks, such as cancer grading[22], finding origins for cancers of unknown primary (CUP)[23] or improved patient stratification[24-26]. However, no previous AI competition for nuclear identification has performed an analysis on how the performance of submitted algorithms impacts downstream applications. We consider this to be particularly important because up until now, there has been limited understanding into what level of performance is required for automatic nuclear identification.

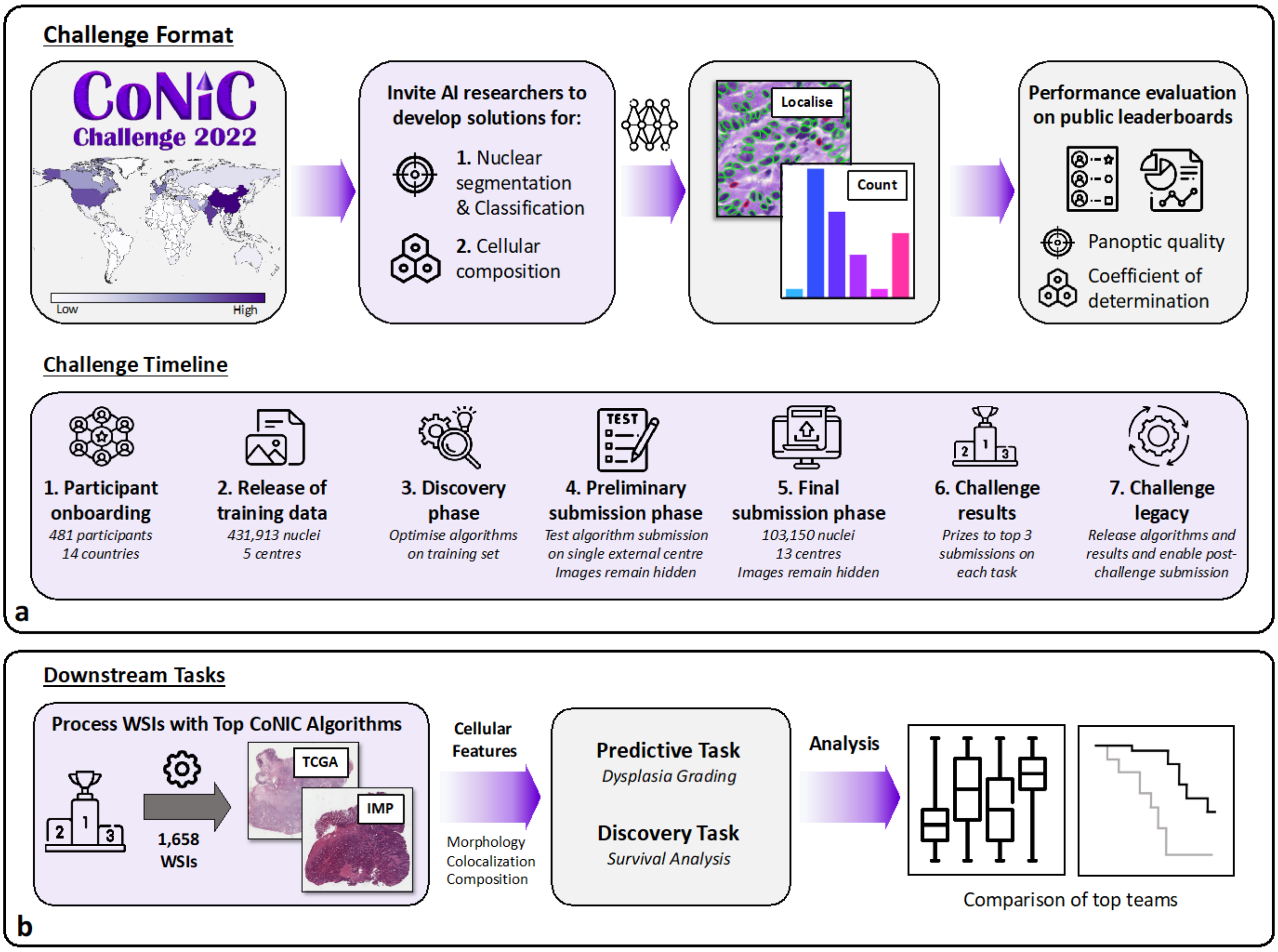

**Fig. 1 | Overview of the CoNIC Challenge. a**, Challenge format and timeline. The format describes the main aims of the challenge, which involves developing AI models for 1) automatic nuclear segmentation and classification and 2) cellular composition. The challenge timeline describes the major events during the competition. **b**, Application of the best performing models from the challenge on downstream tasks. We take the best models from the challenge and assess their performance on the tasks of dysplasia grading and survival analysis.

To counter the above limitations, we organised the Colon Nuclei Identification and Counting (CoNIC) Challenge that invited participants from around the world to develop solutions aimed at solving the following two tasks: 1) nuclear segmentation and classification and 2) prediction of cellular composition. The CoNIC Challenge uses an extension of the current largest available dataset for nuclear instance segmentation and classification, consisting of over 535,000 unique nuclei from 16 centres, which is over 11 times the number of nuclei used in the previous largest challenge. In addition to using a large dataset to ensure reliable evaluation, we also required participants to submit their algorithms, rather than the results, enabling test images to remain hidden and guaranteeing unbiased evaluation.

Furthermore, we performed an extensive assessment of the top algorithms on two clinical tasks: dysplasia grading and survival analysis. For this, we processed 1,658 WSIs from two independent colorectal cohorts with the best performing models and assess the impact of nuclear recognition performance on each downstream application. We identified that the best methods from the challenge are capable of achieving superior performance as compared to previous method. Nevertheless, our findings also suggest that there exist differences in the most important features identified using each of the state-of-the-art model predictions. As a result, subsequent interpretation of these features should be done with caution. To encourage the development of further downstream methods using nuclear features, we also make these WSI-level results, along with the top-performing algorithms, publicly available.

Taking all of this into account, we believe that the CoNIC Challenge will be pivotal in stimulating the development of interpretable cell-based AI models for CPath. The challenge website can be accessed at https://conic-challenge.grand-challenge.org/.

# Results

## CoNIC challenge overview

To stimulate the development of automatic models for nuclear recognition, we organised an AI competition that invited researchers to develop solutions for two tasks: 1) nuclear segmentation and classification and 2) cellular composition. For this purpose, we extended our recent Lizard dataset[14] so that it now contains 535,063 labelled nuclei, making this ten times the size of the previous largest AI competition for automatic nuclear recognition in CPath. Specifically, participants were required to either segment or predict the counts of the following types of nuclei: epithelial, plasma, lymphocyte, neutrophil, eosinophil and connective tissue. Here, we use connective tissue as a broader category consisting of fibroblasts, muscle and endothelial cells.

In Fig. 1a we give an overview of the CoNIC Challenge, including the timeline with the following major events: 1) participant onboarding, 2) release of training data, 3) discovery phase, 4) preliminary submission phase, 5) final submission phase, 6) challenge results, and 7) challenge legacy. The competition was hosted on Grand Challenge (https://grand-challenge.org), which enabled seamless participant registration and provided a platform that allowed algorithm submission. Training data was released at the beginning of the competition, where we extracted small image regions (patches) of size 256×256 from the original Lizard training set and made them available to download.

We provided a HoVer-Net[9] model, which was optimized on the training dataset, as the baseline for the competition. Participants had a two-week preliminary submission phase to familiarise themselves with the submission system and improve their algorithm generalisation. To assist with the latter, we utilised a small sample of the full evaluation dataset, which came from a single TCGA centre with images that looked noticeably different from the training data. The final submission phase lasted one week, where participants could only submit once per task. Within 60 minutes, the algorithms had to process all 103,150 nuclei in the evaluation dataset, which came from images within the Lizard test dataset and a special colon biopsy dataset made for the competition. The results were kept secret until they were revealed at the challenge workshop.

In total, we received 373 submissions during the challenge, where 208 were made for the segmentation and classification task and 165 for the cellular composition task. 26 unique teams appeared on the final segmentation and classification leaderboard, whereas 24 teams appeared on the cellular composition leaderboard. Upon conclusion of the challenge, we have kept the submission portal available and release the top algorithms so that future developments can continue.

## Segmentation and classification results

In Fig. 2a we display the final competition standings for the segmentation and classification task. These are shown in the form of a heatmap, where results are sorted by their final $mPQ^{+}$ score. We observe that the epithelial cell, lymphocyte and connective tissue cell classes were the easiest to segment, with average $PQ^{+}$scores across all participants of 0.513, 0.496 and 0.443, respectively. On the other hand, neutrophil and eosinophil classes were the most difficult nuclei, with average $PQ^{+}$ scores of 0.213 and 0.305. We believe that this was due to the large class imbalance in the dataset, with significantly fewer neutrophils and eosinophils. *EPFL | StarDist*, *MDC Berlin | IFP Bern* and *Pathology AI* were the top three submissions on this task, with $mPQ^{+}$ scores of 0.501, 0.476 and 0.463, respectively. We display visual segmentation and classification results for the top participants in Fig. 3, where we observe that generally submitted models could successfully delineate the boundaries of different nuclei. It was especially impressive to see that submissions such as *EPFL | StarDist* were able to detect neutrophils within the lumen in the 3rd row of the figure. It is evident that some models struggled on the external TCGA dataset. For example, in the 5th row of Fig. 3, *Pathology AI* misclassified plasma cells as epithelial cells, whereas in the bottom row some participants failed to detect various epithelial nuclei.

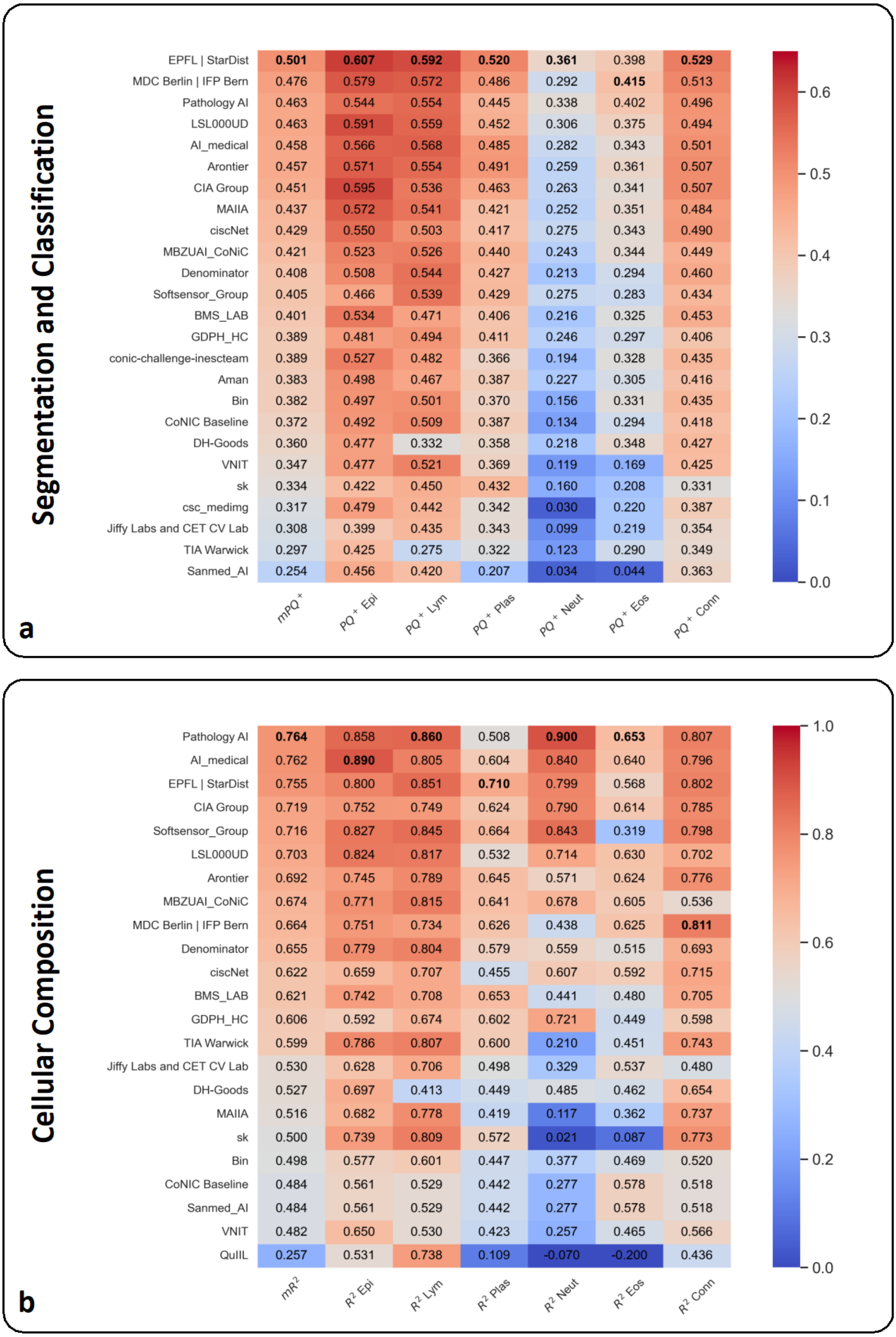

**Fig. 2 | Challenge results on the final test set as heat maps. a**, Final results for the segmentation and classification task. **b**, Results for the cellular composition task. $PQ^+$ and $mPQ^+$ refer to the Panoptic Quality per class and averaged over all classes, respectively. Similarly, $R^2$ and $mR^2$ are the coefficient of determination per class and averaged over all classes.

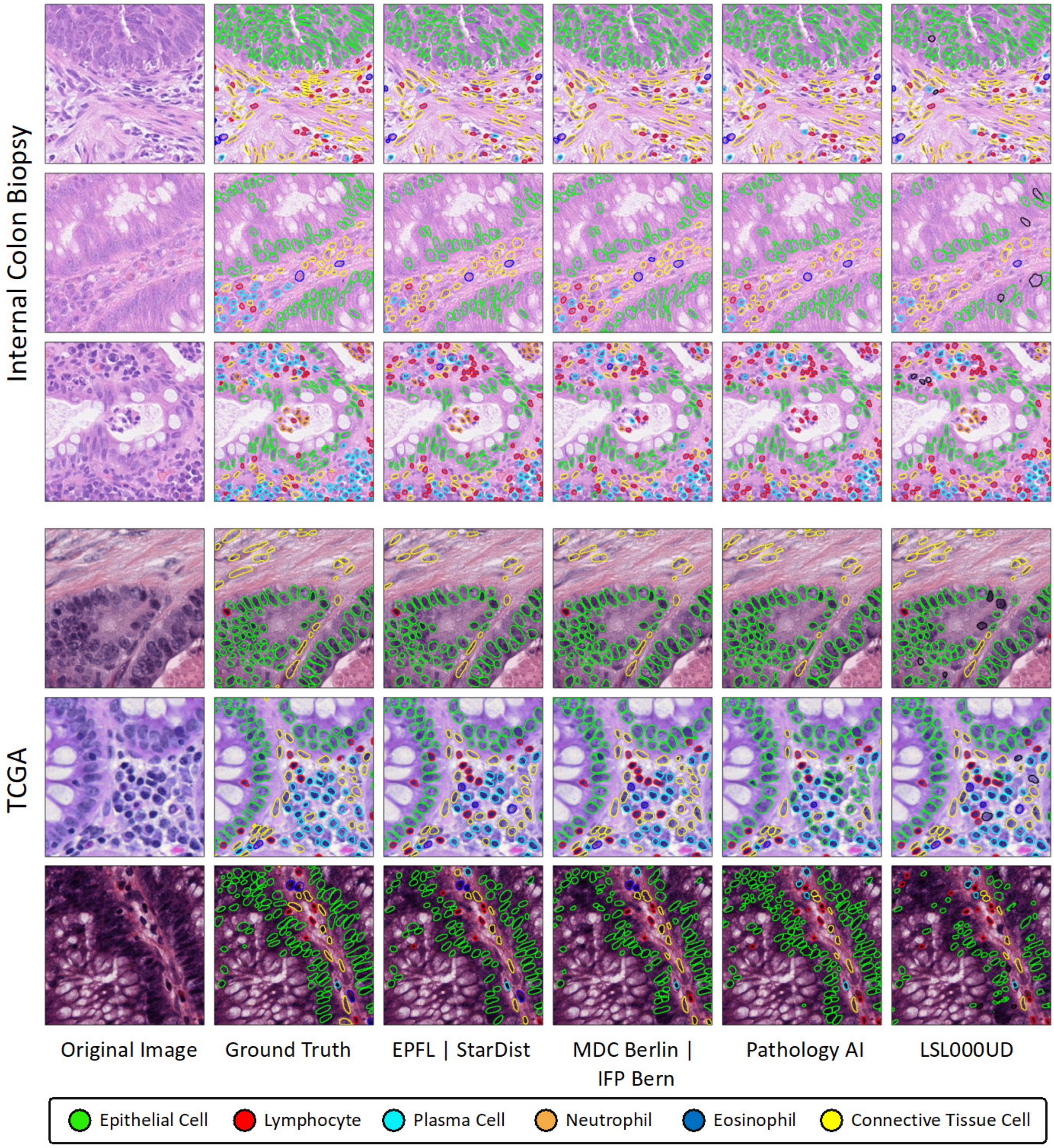

**Fig. 3 | Visual results from the top participants of the segmentation and classification task.** The first 3 rows show results on the internal colon biopsy dataset from UHCW and the bottom 3 rows show results on the TCGA dataset from different submissions as well as the ground truth.

## Cellular composition results

In Fig. 2b, we show the final results of the cellular composition task. Results are sorted in order of their final position on the leaderboard, which was determined by the $mR^2$ score. The standard deviation of the top 20 submissions for $mR^2$ was 0.095, as opposed to 0.042 for $mPQ^+$, indicating that there was greater variability in the results for the cellular composition task. It is evident that participants who were able to sustain a good performance across all classes secured strong positions on the final leaderboard. Again, the epithelial, lymphocyte and connective tissue cell classes were the easiest to predict, with average $R^2$ scores over all submissions of 0.713, 0.722 and 0.673, respectively. The top three submissions for the cellular composition task were *Pathology AI*, *AI_medical* and *EPFL | StarDist* with final scores of 0.7641, 0.7625

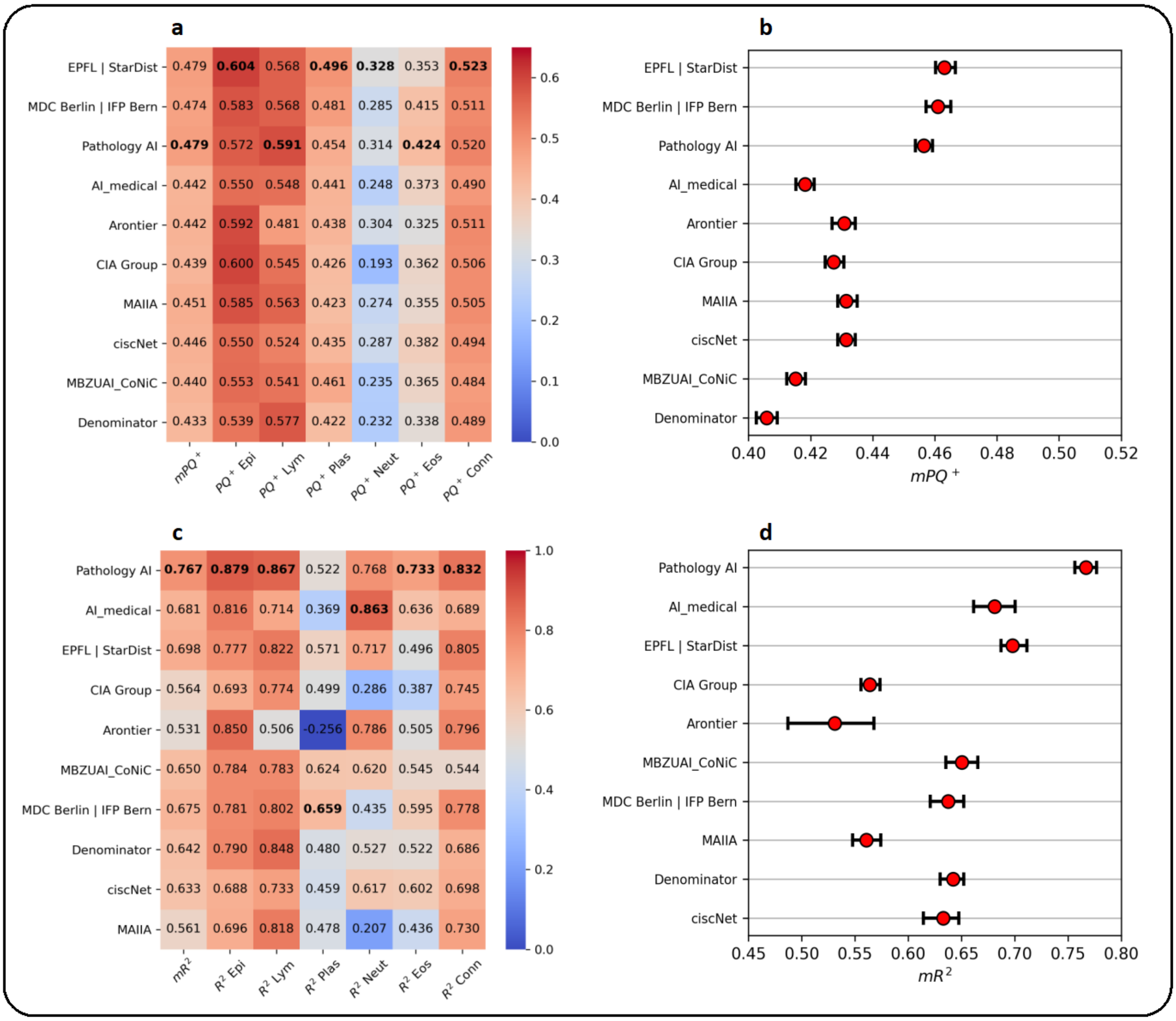

**Fig. 4 | Challenge results of the top participants trained on a single defined split of the data and without model ensembling. a**, **c**, Model results on a single split as a heat map. **b**, **d**, Results using bootstrapping of the test set (n=100) to visualise the confidence bounds. **a**, **b**, Results for segmentation tasks and **c**, **d**, results for cellular composition tasks. $PQ^{+}$ and $mPQ^{+}$ refer to the Panoptic Quality per class and averaged over all classes, respectively. Similarly, $R^2$ and $mR^2$ are the coefficient of determination per class and averaged over all classes.

and 0.7550. Each of these submissions obtained good correlation scores for the minority classes, owing to their strong final positions. As an alternative form of visualisation, we show final results for both tasks as point plots in Extended Data Fig. 1. We also compare the performance with additional metrics for both tasks in Extended Data Fig. 2.

## Impact of model ensembling and bootstrap analysis

To see how the models really perform, we asked the top teams to submit their algorithms that were trained only on a specific data split and without the use of ensembling. Here, ensembling involves combining results from multiple runs of the network with different architectures, checkpoints, or transformed images. We present the results in Fig. 4, with *a* and *b* showing the results for segmentation and classification, and *c* and *d* for cellular composition. Without ensembling, the mean score among these participants decreased from 0.4501 to 0.4330 for segmentation and classification and from 0.6823 to 0.6402 for cellular composition, showing that ensembling has a big impact on the final standing. In Fig. 4*a* and Fig. 4*c*, we show the heatmap of the results, which are ordered by their original ranking in the competition. In parts *b* and *d* of the figure, we perform bootstrapping (n=100) of the submissions for each task to get the confidence bounds. The top three submissions for each task still perform well, even under the conditions we set for this experiment. We show the difference in performance between the original and single model submissions in Extended Data Fig. 3.

### Application of top models to downstream clinical tasks

Accurate recognition of nuclei in histopathology images enables the extraction of interpretable features for downstream clinical pipelines. Therefore, as a next step we assessed the impact of features derived from the top nuclear recognition algorithms on the tasks of dysplasia grading and survival analysis. To enable this, we processed a total of 1,658 WSIs with the top three ranked participants from the segmentation and classification task (*EPFL | StarDist*, *MDC Berlin | IFP Bern* and *Pathology AI*) before extracting a series of global cell-level features. Example visual examples on two randomly selected WSIs are shown in Fig. 5. We only considered models trained on the single split of data because using models with excessive ensembling, like was done in many of the original submissions, is not feasible when processing a large amount of WSIs. For our downstream experiments we used a total of 222 features that can be broadly categorised into the following groups: morphological, density and colocalisation features. These features were directly used as input to a machine learning model for automated diagnosis and patient stratification.

### CoNIC models for cell-based dysplasia grading

To assess the impact of nuclear recognition performance on automated diagnosis, we utilised a dataset of 1,132 colon WSIs from IMP Diagnostics Laboratory in Portugal[28]. All data was H&E-stained and labelled as either non-neoplastic, low-grade dysplasia or high-grade dysplasia. We then used the global nuclear features obtained from the results of each of the top teams as input to a gradient boosted random forest, using five-fold cross validation to ensure reliable results. In Fig. 6*a*, we show the $F_1$ score and Quadratic Weighted Kappa (QWK) over all experimental runs for each team, where we observe that *MDC Berlin | IFP Bern* obtains the best performance with average $mF_1$ and QWK scores of 0.8739 and 0.8463 on the testing set, respectively.

### CoNIC models for cell-based survival analysis

To assess impact of the nuclear features on being able to successfully stratify patients, we predicted overall and disease-specific survival within 526 H&E-stained colorectal WSIs from The Cancer Genome Atlas (TCGA). For this experiment, we utilised the features as input to gradient boosted trees (XGBoost) and performed five-fold cross validation to predict a risk score for each patient. In Fig. 6*b* we show the concordance indices (C-indices) obtained when using the features from each of the top teams for survival tasks. Here, we see that *EPFL | StarDist* obtains the best performance with an average C-index of 0.6554 and 0.6456 respectively for predicting disease specific survival and overall survival. Detailed results for both dysplasia grading and survival analysis can be found in Supplementary Section S2.

## Discussion

The competitive nature of the CoNIC Challenge has enabled rapid development and improvement of models for automatic nuclear recognition, leading to many final submissions surpassing the previous state-of-the-art[9]. Most participants used DL models with an encoder-decoder architecture and particularly successful submissions employed a strategy to deal with the significant class imbalance present in the dataset, such as patch oversampling or weighted loss functions. In fact, many successful submissions developed a subtly modified version of HoVer-Net, suggesting that our provided baseline provided a good foundation for participants to build upon. We saw that the baseline HoVer-Net approach could be improved by considering a stronger backbone or instance segmentation target such as with additional directional distance maps. We provide a visual summary of the algorithms submitted by the participants, along with the training details in Extended Data Fig. 13. We also provide a more detailed description of each of the approaches in Supplementary Section S1.

Despite us allowing participants to treat each task independently, nearly all submissions inferred the cellular composition from the segmentation and classification output. However, those teams that predicted the cellular composition directly from the original image did not achieve a high ranking on the final leaderboard. Determination of the cellular composition allows us to effectively model the tumor microenvironment (TME) of the tissue, which has been shown to be particularly powerful as a prognostic indicator. With the challenge acting as a facilitator to improve automatic nuclear profiling, we hope that it will stimulate the development of advanced methods correlating the TME to patient outcome.

Overall, we found that participants were able to achieve a strong performance on both tasks on our developed dataset. However, despite significant advancements being made within the challenge, additional work is required to further boost the ability to recognise minority classes, such as neutrophils and eosinophils. This may be achieved with the help of additional data and the development of new strategies for dealing with the class imbalance. A particularly interesting technique that was used in the challenge (*Arontier* and *Aman*) is copy-and-paste augmentation, which can be used to artificially increase the number of under-represented nuclei in the dataset. Another strategy that appears promising is the utilisation of generative methods to create synthetic images containing minority classes, while preserving the expected spatial configuration of nuclei within the tissue[29].

The dataset that we introduced as part of the challenge is the largest existing dataset of nuclear segmentation and classification in CPath. Despite this, our dataset is currently from a single tissue type and so we cannot guarantee that models developed during the challenge will generalise to unseen tissues, despite various inflammatory cells appearing the same across different organs. In future work, we may extend the current dataset to other major tissue types, such as breast, prostate and lung to increase the range of downstream applications that the models can be applied to. Also, we currently group endothelial cells, fibroblasts and muscle cells into a single category. Explicit separation of these classes will enable the consideration of features such as cancer-associated fibroblasts and endothelial cell morphology, which can be prognostically informative[6,7,30]. We are also aware of the limitations that exist as a result of labelling nuclei using only routine H&E slides. In future work, perhaps it would be advantageous to instead rely on co-registered H&E and IHC slides to provide more accurate ground truth, especially for immune cell subtyping.

We ensured that our dataset was sufficiently large to provide a good indication for how models perform across a range of scanner types and lab preparation methods. However, this does not guarantee that developed models will work out-of-the-box when deployed in a clinical setting. Future work may include a thorough investigation into the robustness of AI models for nuclear identification[31,32], where lessons learned can help reduce the likelihood of unexpected model behaviour in the *wild*.

In addition to the main challenge, we utilised the baseline and three best performing models from the segmentation and classification task and processed 1,658 WSIs from two datasets, with the intention of understanding how the performance of automated nuclear identification affects downstream clinical tasks and identifying whether state-of-the-art (SoTA) methods can improve upon the baseline. In particular, we extracted patient-level features from the WSI results and used them as input to perform automatic dysplasia grading and survival analysis. From Fig. 6, the digital features based on the SoTA methods that were identified from the CoNIC challenge are all significantly more predictive compared to those from the baseline ($p \ll 0.001$, student-t test). This suggests that accurate models for nuclear recognition may in fact be essential when using associated features for downstream tasks. Although there is an apparent relationship between CoNIC results and the performance of subsequent tasks, further work is required to understand the implications of further significant boosts in nuclear recognition. We found that despite small differences in the best results obtained within the challenge, there was notable variation in the importance of features when applied to downstream tasks. Given that many studies typically utilise predictions from a single nucleus segmentation and classification method, our results raise questions on the validity of identified digital features using just a single approach, especially for survival analysis problems.

Specifically, when grading dysplasia, while all three SoTA methods achieved similar performance, compared to the other two methods, *MDC Berlin | IFP Bern* disproportionately utilised statistics concerning plasma nuclei morphology while neglecting those from eosinophils (Extended Data Fig. 6a). On the other hand, neutrophil morphology was determined more important for *EPFL | StarDist* than any other method. In spite of these differences, all three methods considered the morphology of epithelial nuclei as important features for accurately predicting dysplasia grade (Extended Data Fig. 7-9). This finding aligns with existing clinical observations[33].

For survival analyses, from Fig. 5b and 5c and Extended Data Fig. 5 and 6, we observe that SoTA methods not only have significant variation in performance but also have a different set of identified features for a given task. Despite this, we found that statistics concerning the TME of neutrophils were consistently identified as important features, as shown in Extended Data Fig. 7 and 10-12. This observation aligns with existing clinical studies[34] which perhaps encourages further investigation into neutrophils and their impact and emphasizes the need to further improve neutrophil detection performance. Also, from Extended Data Fig. 7 and 10-12, while not as informative as neutrophil-based features, all teams also agree that a higher cellular composition of eosinophils relate to better patient outcome. This observation confirms a recent clinical finding[35] which states that eosinophils have anti-tumourigenic properties in colorectal cancers.

As a result of its competitive nature, the utilisation of the largest dataset of its kind and the existence of a rigorous evaluation protocol, we believe that the CoNIC Challenge has been largely influential in helping to further push forward the state-of-the-art for automatic nuclear recognition in CPath. To foster the development of future approaches for cell-based biomarker exploration, we are releasing the WSI-level results using the best methods from the challenge. We are also accepting post-challenge submissions using the same evaluation framework as the original competition.

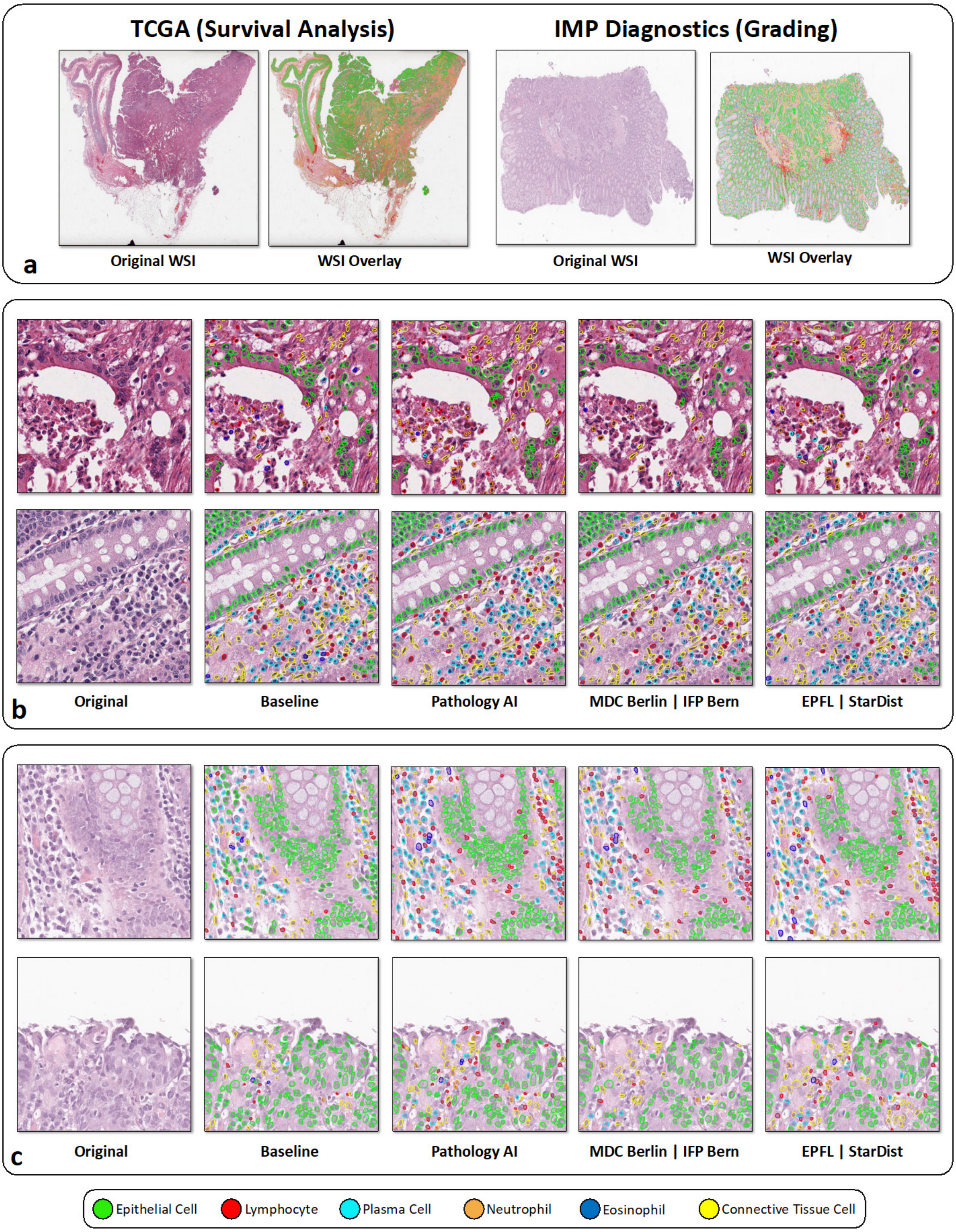

**Fig. 5 | WSI-level visual results from the top submissions on the segmentation and classification task trained on a single pre-defined split of the data. a**, Original WSI along with an example nuclear segmentation overlay. **b**, Zoomed-in predictions of the top three teams compared to the baseline for the example TCGA WSI. **c**, Zoomed-in predictions of the top three teams compared to the baseline for the example IMP Diagnostics WSI. For **b** and **c**, each row shows a different region from the same WSI.

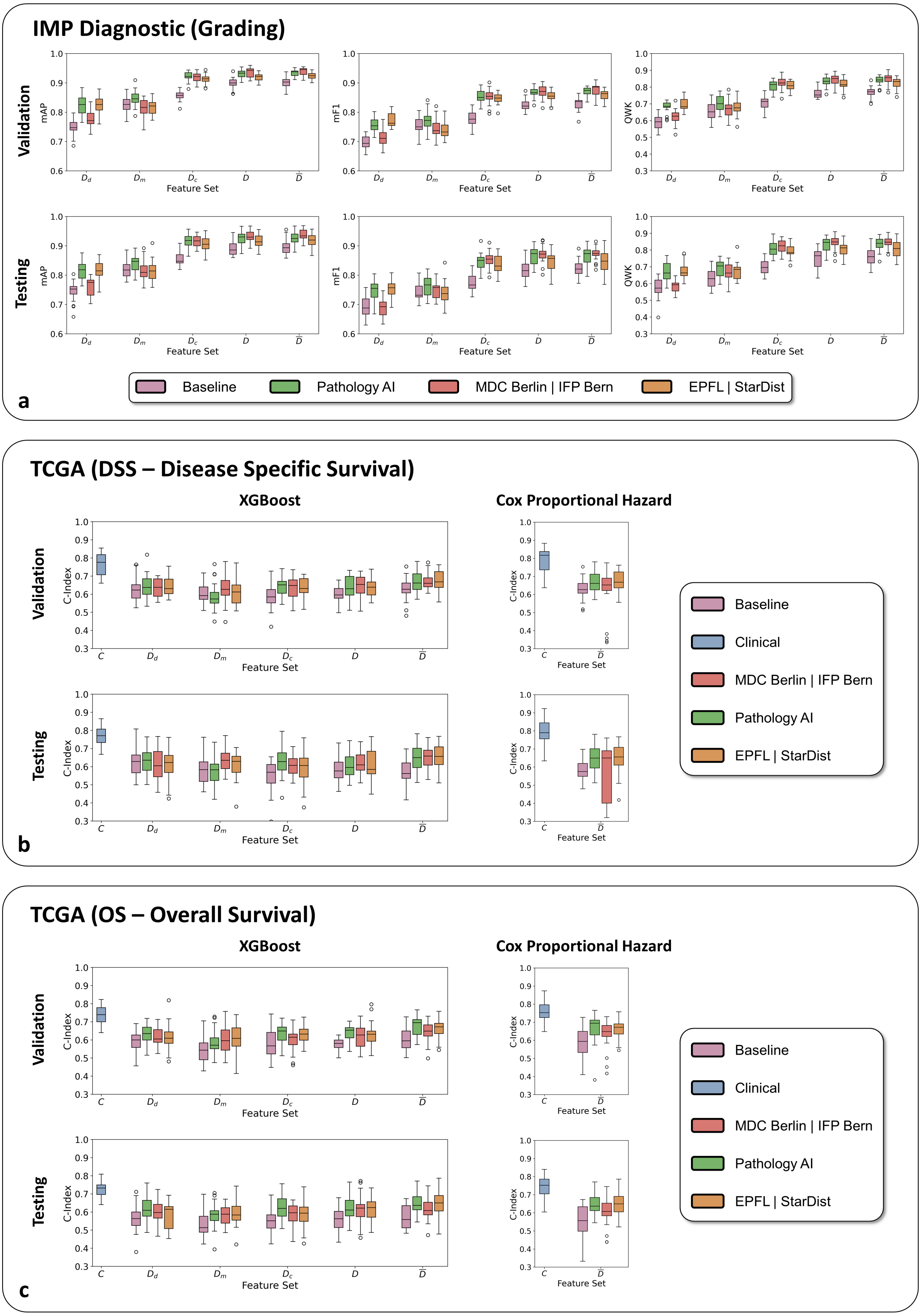

**Fig. 6 | Results when using nucleus features computed based on the predictions of the top three teams in segmentation and classification task for patient-level grading and survival analysis. a**, Results for grading, **b**, results for disease-specific survival analysis and **c**, results for overall survival analysis. $mF_1$ and $mAP$ denote the mean $F_1$ and mean Average Precision scores, respectively. C-Index is the concordance index, which is a commonly used metric for survival analysis. $D_d$, $D_m$ and $D_c$ refer to density-based, morphological and colocalization features, respectively. $D$ is the entire feature set and $\bar{D}$ is the set after feature selection. $C$ is the set of clinical features.

# Methods

## Challenge design

The challenge began on the 20th November 2021, when we released the training data, so that participants could start developing solutions for the two tasks. We also released evaluation code and a HoVer-Net[9] baseline model on our challenge GitHub page to help accelerate model development and prevent participants needing to build models from scratch (https://github.com/TissueImageAnalytics/CoNIC). During this time, participants were able to ask questions on the webpage forum to further understand the intricacies of the tasks and the baseline model.

In the meantime, we implemented an evaluation framework that enabled participants to submit their algorithms to the competition, allowing us to keep test images hidden and ensuring unbiased evaluation. For this, we utilised the Grand Challenge platform (https://grand-challenge.org) developed by the Diagnostic Image Analysis Group at Radboud University Medical Center. Participants were required to submit their algorithms to the portal as Docker containers, which were then used to process the test images in the cloud using Amazon Web Services (AWS). To help with this, we created detailed videos on the challenge webpage (https://conic-challenge.grand-challenge.org/), providing step-by-step instructions on how to create, test and submit the containers to the portal. To facilitate this, we provided a Docker template, along with a specific example using our baseline, that participants could easily adapt for their own solutions. Adhering to this template ensured that containers were able to appropriately read image data in the backend, process the data using their developed algorithms and return outputs easily recognisable in the next step of the evaluation protocol.

For evaluation, we developed a container that took the algorithm outputs and computed the metrics of each submission. These metrics then interacted with the Grand Challenge platform, where the overall results were then displayed on the public leaderboards. This overall submission procedure could be tested during the preliminary submission period, that took place between 13th-27th February 2022. At this stage, the organisers regularly contacted with competing teams, advising them on potential reasons for failed submissions. Therefore, not only did this encourage the improvement in the performance of developed models due to its competitive nature, but it also prepared participants for making their final submissions. Teams were allowed one submission per day for each task, leading to many results being displayed on the leaderboards. For each task, a separate submission was required, even if the segmentation output was used to predict cellular composition. However, it was not mandatory to complete both tasks and participants could focus on just one, such as predicting cellular composition. This has recently been done by Dawood *et al.*[36] where the counts of different cell types were predicted without explicitly localising each nucleus. The final stage was between 27th February-6th March 2022, where only one successful submission was permitted for each team per task. Here, we allowed a maximum of 60 minutes to process the full test set, to prevent excessive ensembling.
Upon conclusion of the challenge, we invited the top ten teams to send us two Docker containers: 1) original submission and 2) models trained on a specified split of the data, without ensembling. Original algorithms were requested so that we could make them available to the public (for those that granted us permission), whereas retrained models on a single split enabled us to perform a fairer head-to-head comparison between methods. Gathering models that did not use ensembling also unlocked the potential to use them for WSI processing and downstream analysis, due to reasonable inference times. Docker containers are available for download by visiting https://warwick.ac.uk/conic-challenge. Despite the conclusion of the challenge, participants can still submit algorithms and visualise their results on post-challenge leaderboards using the same Docker-based submission protocol as outlined above.

## CoNIC challenge dataset

To ensure reliable evaluation and to foster the development of generalisable models, it is essential for AI competitions to utilise large datasets. Therefore, in this competition, we used data from our recently curated Lizard dataset[14], consisting of 495,179 nuclei in H&E-stained image regions from 16 different centres and three countries. This is approximately ten times the number of nuclei considered in the previous largest competition for nuclear recognition in CPath[21]. To gather such a large dataset, we employed an iterative approach to annotate the data, which used a combination of semi-automatic and manual refinement steps with significant pathologist involvement. Utilising this strategy ensured the development of an accurate dataset at scale, resulting in the largest available dataset for nuclear segmentation and classification in CPath. Nuclei were labelled in accordance with their associated cell type and categorised as either: epithelial cell, lymphocyte, plasma cell, neutrophil, eosinophil or connective tissue cell. Here the connective tissue cell category groups endothelial cells, fibroblasts and muscle cells into a single class. Therefore, models trained on this data may be used to help effectively profile the colonic tumour micro-environment. We choose to focus on nuclei from colon tissue to ensure that our dataset contains images from a wide variety of different normal, inflammatory, dysplastic and cancerous conditions in the colon - therefore increasing the likelihood of generalisation to unseen examples.

In addition to the Lizard dataset[14], we labelled 39,884 nuclei from an internal colon biopsy dataset. This was done by multiple pathologists in the form of point annotations with consensus review[37]. Then, segmentation masks were produced using a semi-automatic method[13] and results were manually refined. This led to a total challenge dataset of 535,063 accurately labelled nuclei. For the purpose of the competition, we then extracted patches of size 256×256 pixels at 20× objective magnification (approximately 0.5 microns/pixel) and provided the counts within the central 224×224 pixel region. This ensured that nuclei were only considered if the majority of pixels were visible. We provide a detailed summary of the breakdown of the dataset in Extended Data Fig. 4.

## Challenge evaluation metrics

For each task, we utilised a single metric that was used to rank team submissions. For the segmentation and classification task, the multi-class panoptic quality was used, which has recently been justified as a strong metric by Graham *et al.*[9]. Here, for each type $t$, the $PQ$ is defined as:

$$PQ_t = \frac{|TP_t|}{|TP_t| + \frac{1}{2}|FP_t| + \frac{1}{2}|FN_T|} \times \frac{\sum_{(x_t, y_t) \in TP} IoU(x_t, y_t)}{|TP_t|},$$

where $x$ denotes a ground truth (GT) instance, $y$ denotes a predicted instance, and $IoU$ denotes intersection over union. Setting $IoU > 0.5$ will uniquely match $x$ and $y$. This unique matching therefore splits all available instances of type $t$ within the dataset into matched pairs ($TP$), unmatched GT instances ($FN$) and unmatched predicted instances ($FP$). Henceforth, we define the multi-class $PQ$ ($mPQ$) as the task ranking metric, which takes averages the $PQ$ over all classes. Note, for $mPQ$ we calculate the statistics over all images to ensure there are no issues when a particular class is not present in a patch. This is different to $mPQ$ calculation used in previous publications, such as PanNuke[12], MoNuSAC[21] and in the original Lizard paper[14], where the $PQ$ is calculated for each image and for each class before the average is taken. Hence, for the purpose of this challenge, we refer to the metric as $mPQ^+$. As an added benefit, $PQ$ can be easily decomposed into Detection Quality ($DQ$) and Segmentation Quality ($SQ$), enabling a more detailed analysis of participant's results. Despite these results not being utilised in the main challenge, we display a summary of the obtained $mDQ^+$ and $mSQ^+$ scores obtained for each team in Extended Data Fig. 2.

For the cellular composition task, we used the multi-class coefficient of determination to determine the correlation between the predicted and true counts. Similar to the previously described metric, the statistic is calculated for each class independently and then the results are averaged. In particular, for each nuclear category $t$, the correlation of determination is defined as follows:

$$R_t^2 = 1 - \frac{RSS_t}{TSS_t},$$

where $RSS$ stands for the sum of squares of residuals and $TSS$ stands for the total sum of squares after fitting a regression line is fitted to the predicted and actual counts. For additional analysis, we also display additional regression results, using Mean Absolute Error (MAE) and Mean Arctangent Absolute Percentage Error (MAAPE), in Extended Data Fig. 2. As before, both metrics compute the statistics for each class independently and the results are then averaged to give the final score. Unlike other metrics described throughout this paper, low scores for MAE and MAAPE indicate a strong performance.

## Downstream clinical tasks

### Clinical datasets

To assess how differences in the results of nuclear detection across various teams impacts the performance of downstream tasks, we collected data containing Haematoxylin and Eosin (H&E) stained WSIs of colorectal tissue from The Cancer Genome Atlas (TCGA) and IMP Diagnostics Laboratory. Here, the TCGA dataset was used to perform survival analysis, whereas the IMP Diagnostics dataset was used for dysplasia grading. TCGA slides were digitised at various institutions and therefore the scan resolution of the original WSIs varies. Slides from IMP Diagnostics were digitised with a Leica GT450 scanner at a pixel resolution of 0.263 MPP. In total, we obtained 526 WSIs of surgical resections from TCGA and 1,132 WSIs of endoscopic biopsies from IMP Diagnostics.

To enable survival analysis on TCGA, we extracted the disease-specific and overall survival times as well as the respective survival statuses of each patient. Within the IMP Diagnostics dataset, each slide is categorised into one of the following groups: non-neoplastic, low-grade lesion or high-grade lesion. Here, non-neoplastic slides contained both normal and inflammatory conditions, low-grade lesions contained conventional adenomas with low-grade dysplasia, and high-grade lesions contained conventional adenomas with high-grade dysplasia, intra-mucosal carcinomas and invasive adenocarcinomas.

To reliably evaluate the performance of each downstream task, on both datasets we performed five-fold cross-validation. Each fold was separated into a training (60%), validation (20%) and test set (20%). We repeated this procedure five times with different random seeds, resulting in a total of 25 different splits of the data.

### Evaluation metrics

For dysplasia grading, we measured the $F_1$ and Average Precision ($AP$) score per category and then calculated the average result, denoted by $mF_1$ and $mAP$. In addition, we computed the Quadratic Weighted Kappa (QWK), which measures the agreement between the predictions and true diagnostic categories. For survival analysis, we measured and reported the concordance index (C-Index) between the predicted risk scores and actual events.

### Digital features

For both cohorts, we processed slides using the top-performing models from the segmentation and classification task (*EPFL | StarDist*, *MDC Berlin | IFP Bern* and *Pathology AI*) that were trained on a single split of the challenge dataset. For each WSI, we then extracted 222 patient-level features that could be grouped into the following categories: morphological, colocalisation, and density features. Here, morphological features included the best alignment metric[38] (BAM), size, eccentricity, major axis, minor axis and perimeter of each nucleus based on its predicted contour. Colocalisation features describe the spatial relationship between different types within several pre-defined neighbourhoods (200$\mu$m and 400$\mu$m radius[39,40]). For morphological and colocalisation features, we first calculated these statistics per nucleus and then computed the mean and standard deviation across all WSIs belonging to each patient to give the corresponding patient-level features. Density features describe the global ratio of different nucleus types across all tissue samples of a patient. Depending on the settings of subsequent experiments, a patient-level digital descriptor for a WSI can either contain only morphological, colocalisation, density features, or contain a combination of them all. For clarity, we respectively denoted these digital feature sets as $D_m$, $D_c$, $D_d$ and $D$. A comprehensive description of these features is provided in the supplementary material.

### Predictors for downstream tasks

In order to identify how the nuclear segmentation results for each team affect downstream tasks, it is important for us to not only use a model with a strong predictive power, but also use one that allows interpretation of which input features are important. With this in mind, we utilised a tree-based method, named gradient boosted trees (implemented with XGBoost[41]), for both grading and survival analysis tasks. Compared to other tree-based implementations, XGBoost is well-known for being computationally efficient and scalable, while still ensuring a strong performance[41]. Throughout the paper, we used Random Search on the XGBoost parameter space, with 2048 sampled points, to obtain the best parameters for each set of input features that were obtained from each team and for each of the downstream tasks. The XGBoost parameter sets that have the best validation results across all folds and repetitions are selected for subsequent feature interpretation and analyses on their corresponding testing set.

### Feature selection

As commonly described in other works[42,43] not all input features will necessarily contribute to the considered downstream tasks. Thus, we perform feature selection on $D$ to find the most predictive feature set $\overline{D}$ for the final models. To identify these features, we first performed the previously described random parameter search, but rather than selecting the model that performs best across all folds and repetitions, we selected the best model on each fold, resulting in 25 different models. For each of these 25 models, we examined the impact of the input feature set $D$ on the results (QWK for grading and C-index for survival analysis) using the Permutation Test[44], which gives an importance value for each feature. Then, we averaged the feature importance values across all folds to obtain the overall importance of the features in $D$. Finally, we selected features whose importance scores were greater than the median value across the 222 features. A rough summary of which features were selected from each team and for each task is reported in Extended Data Fig. 5.

### Dysplasia grading

For this task, we extracted $D_m$, $D_c$, $D_d$, $D$ and $\overline{D}$ from the IMP Diagnostics dataset, denoting the various feature sets, as outlined above. Then, using the previously described procedure, we fit XGBoost models on each feature set extracted from the nuclear segmentation results of each team and performed a comprehensive comparative analysis.

### Survival analysis

For this task, as well as comparing the performance of each digital feature set, we also assess how the final selected set of digital features $\overline{D}$ compares with existing clinical features for predicting disease-specific and overall survival. In this work, we utilised sex, age and cancer stage, denoted by $C$, as the set of clinical features. Similar to the grading pipeline, we extracted $D_m$, $D_c$, $D_d$, $D$ and $\overline{D}$ feature sets from the TCGA dataset for utilisation in the downstream survival analysis pipeline. To perform survival analysis, we evaluated the predicted risk scores obtained from fitting XGBoost models on the different feature sets from each team and reported their C-Index on the validation and testing sets.

## Data availability

The development set used as part of this challenge can be downloaded at https://conic-challenge.grand-challenge.org/Data/, where the data is held under a non-commercial Creative Commons license. We provide information on how to download method description papers, team algorithms and WSI-level results at https://warwick.ac.uk/conic-challenge. Information regarding the IMP cohort can be found in the original publication[28]. TCGA WSIs can be downloaded at https://portal.gdc.cancer.gov/.

## Code availability

Evaluation code used within the challenge, along with example notebooks can be found at the following repository: https://github.com/TissueImageAnalytics/CoNIC. A template for making code-based challenge submissions can be found in a separate branch of the same repository. Code used for running the baseline can be found in a separate branch of the original HoVer-Net repository at https://github.com/vqdang/hover_net/tree/conic.

## Acknowledgements

SG, MJ, DS, SR, FM and NR would like to acknowledge the support from the PathLAKE digital pathology consortium which is funded by the Data to Early Diagnosis and Precision Medicine strand of the government's Industrial Strategy Challenge Fund, managed and delivered by UK Research and Innovation (UKRI). FM acknowledges funding from EPSRC grant EP/W02909X/1. We thank Georgios Hadjigeorghiou and Thomas Leech for initial discussions regarding the challenge setup.

## Author contributions

SG, QDV, MJ, SEAR, FM and NR designed and conducted the study. SG curated the challenge dataset. QDV set up the docker-based submission and evaluation framework for the challenge. SG and QDV performed analysis and interpretation of the results. QDV processed all slides with the best challenge algorithms. SG and QDV analysed the impact of WSI-level nuclear recognition results on downstream applications. DS, SEAR, FM and NR provided technical and material support. SG wrote the first draft of the paper. All authors read and approved the final paper.

## Competing interests

SG, DS and NR are co-founders of Histofy Ltd. DS reports personal fees from Royal Philips, outside the submitted work. NR and FM report research funding from GlaxoSmithKline.

## The CoNIC Challenge Consortium[◊]

Dagmar Kainmueller[1,2], Carola-Bibiane Schönlieb[3], Shuolin Liu[4], Dhairya Talsania[5,6], Yughender Meda[5,6], Prakash Mishra[5,6], Muhammad Ridzuan[7], Oliver Neumann[8], Marcel P. Schilling[8], Markus Reischl[8], Ralf Mikut[8], Banban Huang[9], Hsiang-Chin Chien[10], Ching-Ping Wang[10], Chia-Yen Lee[11], Hong-Kun Lin[12], Zaiyi Liu[13], Xipeng Pan[13], Chu Han[13], Jijun Cheng[14], Muhammad Dawood[15], Srijay Deshpande[15], Raja Muhammad Saad Bashir[15], Adam Shephard[15], Pedro Costa[16,17], João D. Nunes[16,17], Aurélio Campilho[16,17], Jaime S. Cardoso[16,17], Hrishikesh P S[18], Densen Puthussery[18], Devika R G[19], Jiji C V[19], Ye Zhang[20], Zijie Fang[21], Zhifan Lin[20], Yongbing Zhang[20], Chunhui Lin, Liukun Zhang[22], Lijian Mao[22], Min Wu[22], Vi Thi-Tuong Vo[23], Soo-Hyung Kim[23], Taebum Lee[24], Satoshi Kondo[25], Satoshi Kasai[26], Pranay Dumbhare[27], Vedant Phuse[27], Yash Dubey[27], Ankush Jamthikar[27], Trinh Thi Le Vuong[28], Jin Tae Kwak[28], Dorsa Ziaei[29], Hyun Jung[29], Tianyi Miao[29]

[1]Max-Delbrueck-Center for Molecular Medicine in the Helmholtz Association, Berlin, Germany
[2]Humboldt University of Berlin, Faculty of Mathematics and Natural Sciences, Berlin, Germany
[3]Department of Applied Mathematics and Theoretical Physics, University of Cambridge, United Kingdom
[4]Department of Electrical Engineering and Automation, AnHui University, HeiFei, China
[5]Softsensor.ai, Bridgewater, New Jersey, United States of America
[6]PRR.ai, Texas, United States of America
[7]Computer Vision Department, Mohamed Bin Zayed University of Artificial Intelligence, Abu Dhabi, United Arab Emirates
[8]Institute for Automation and Applied Informatics Karlsruhe Institute of Technology Eggenstein-Leopoldshafen, Germany
[9]School of Computer Science and Cyber Engineering Guangzhou University, Guangzhou, China
[10]Department of Electrical and Computer Engineering, National Yang Ming Chiao Tung University, Hsinchu, Taiwan
[11]Department of Electrical Engineering, National United University, Miaoli, Taiwan
[12]Institute of Biomedical Engineering, National Yang Ming Chiao Tung University, Hsinchu, Taiwan
[13]Department of Radiology, Guangdong Provincial People's Hospital, Guangdong Academy of Medical Sciences, Guangzhou, Guangdong, China
[14]School of Computer Science and Information Security, Guilin University of Electronic Technology, Guilin, Guangxi, China
[15]Tissue Image Analytics Centre, University of Warwick, Coventry, United Kingdom
[16]Faculty of Engineering, University of Porto, Porto, Portugal
[17]Institute for Systems and Computer Engineering, Technology and Science, Porto, Portugal
[18]FMS Lab, Founding Minds Software Cochin, India
[19]Department of Electronics and Communication College of Engineering Trivandrum, India
[20]Shenzhen Graduate School, Harbin Institute of Technology University Shenzhen, China
[21]Tsinghua Shenzhen International Graduate School, Shenzhen, China
[22]Research and Development center Zhejiang Dahua Technology Co., Ltd Hangzhou, Zhejiang, China
[23]Department of AI Convergence Chonnam National University Gwangju, South of Korea
[24]Department of Pathology, Chonnam National University Medical School Gwangju, South of Korea
[25]Muroran Institute of Technology, Hokkaido, Japan
[26]Niigata University of Healthcare and Welfare, Niigata, Japan
[27]Visvesvaraya National Institute of Technology, Nagpur, India
[28]School of Electrical Engineering, Korea University, Seoul, Republic of Korea
[29]Advanced Biomedical Computational Science, Frederick National Laboratory for Cancer Research, Frederick, United States of America

[◊]The consortium comprises of an extended list of authors contributing towards the challenge.
Consortium contact: jkwak@korea.ac.kr

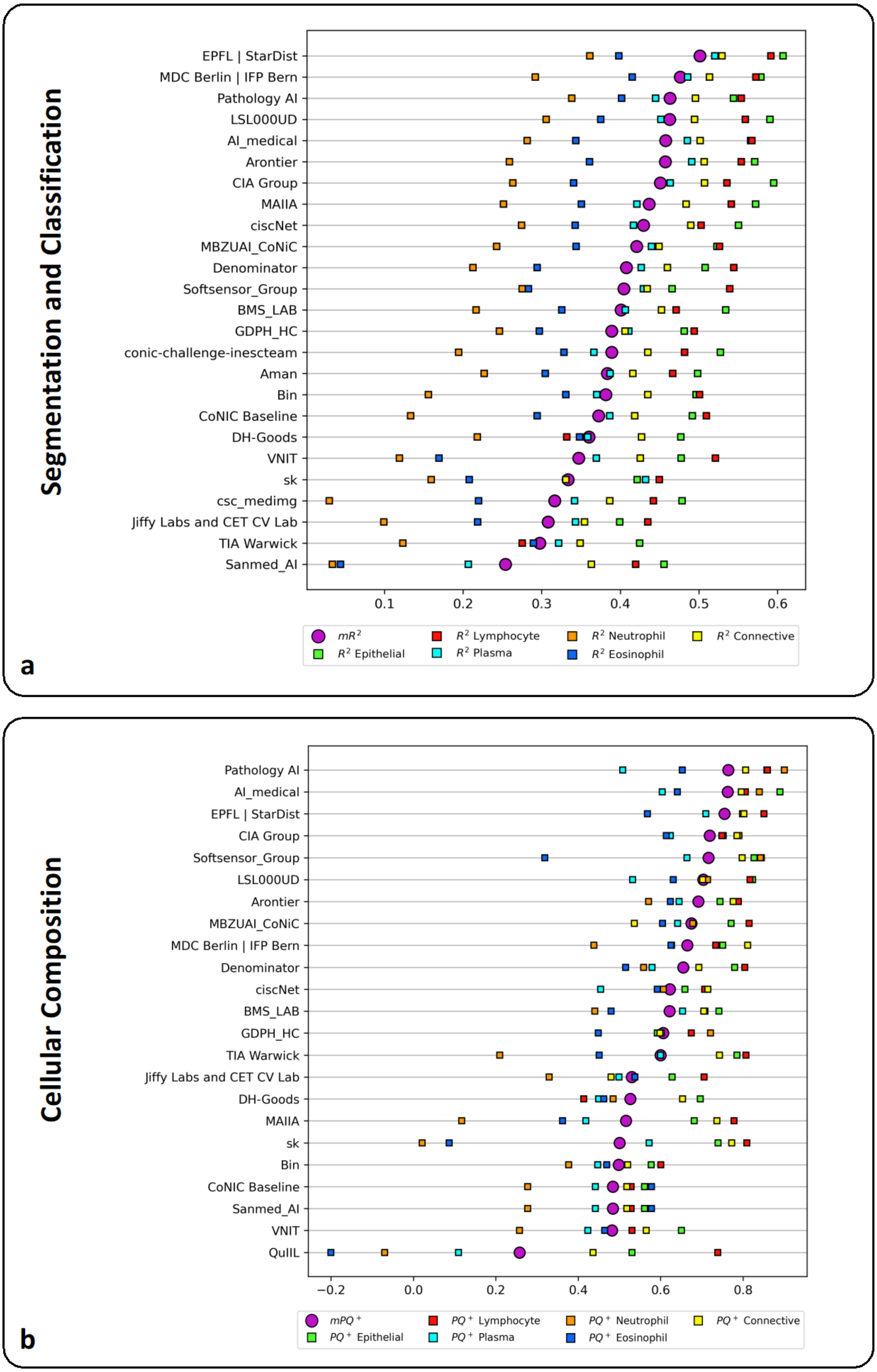

**Extended Data Fig. 1 | Final challenge results for both tasks.** These results are the same provided in Fig. 2, but are shown as point plots as an alternative way of visualising the results.

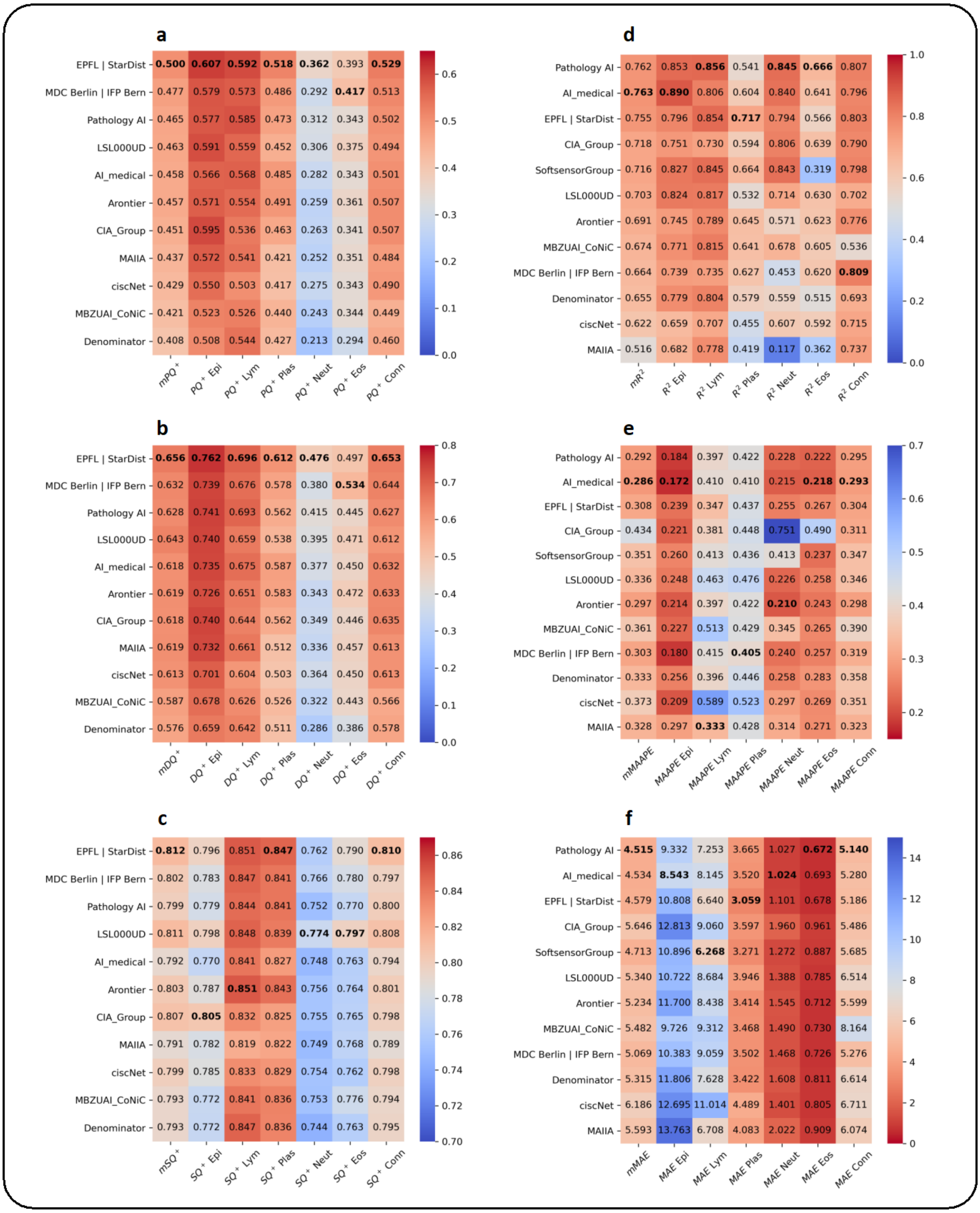

**Extended Data Fig. 2 | Additional results for each task using alternative metrics.** We computed these results using the algorithms that were sent by the participants, which explains slight differences in the results compared to the original standings. The left side (**a**, **b**, **c**) show segmentation and classification results. **a**, $mPQ^+$, **b**, $mDQ^+$ and **c**, $mSQ^+$. The right side (**d**, **e**, **f**) show cellular composition results. **d**, $mR^2$, **e**, $mMAE$ and **f**, $mMAAPE$.

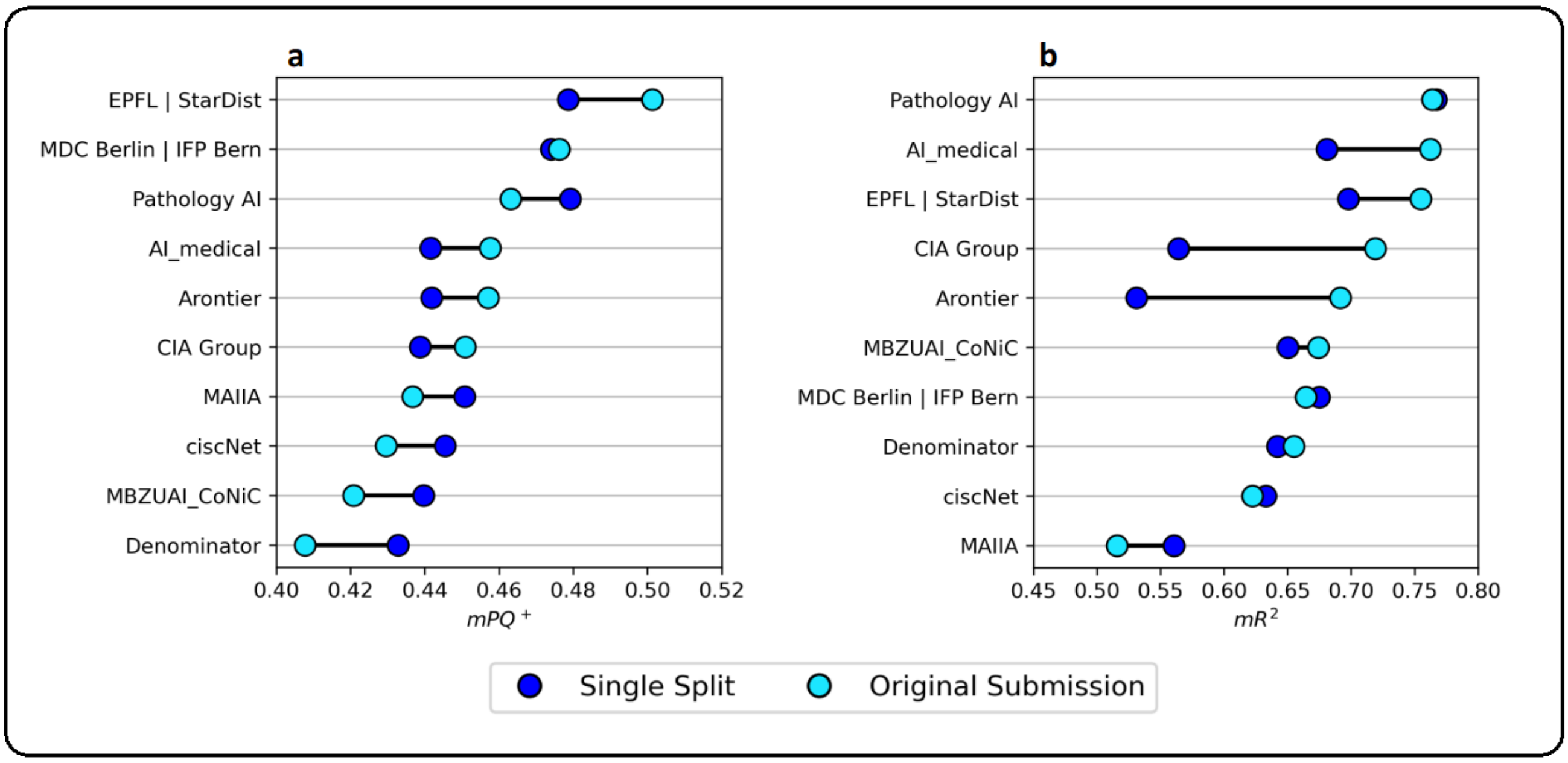

**Extended Data Fig. 3 | Difference in results of original submission compared to those obtained using the models trained on a single split of the data and without ensembling.** **a**, Segmentation and classification results, in terms of $mPQ^+$ and **b**, cellular composition results, in terms of $mR^2$.

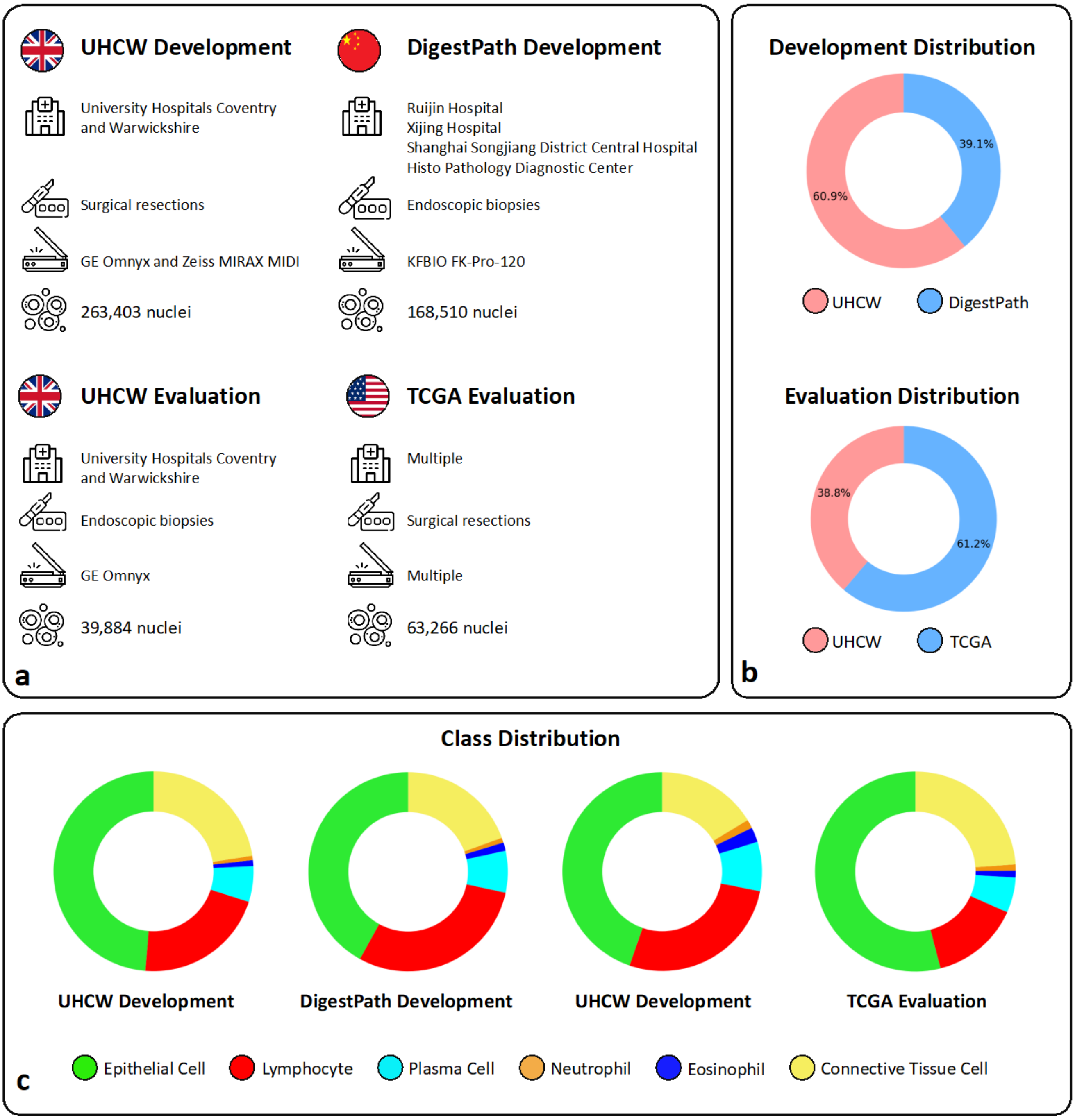

**Extended Data Fig. 4 | Summary of the datasets used in the challenge**. **a**, Information regarding the institution(s), specimen type, scanner manufacturer and number of labelled nuclei. **b**, Distribution of data in the development and evaluation sets. **c**, Distribution of nuclear classes across the different data subsets.

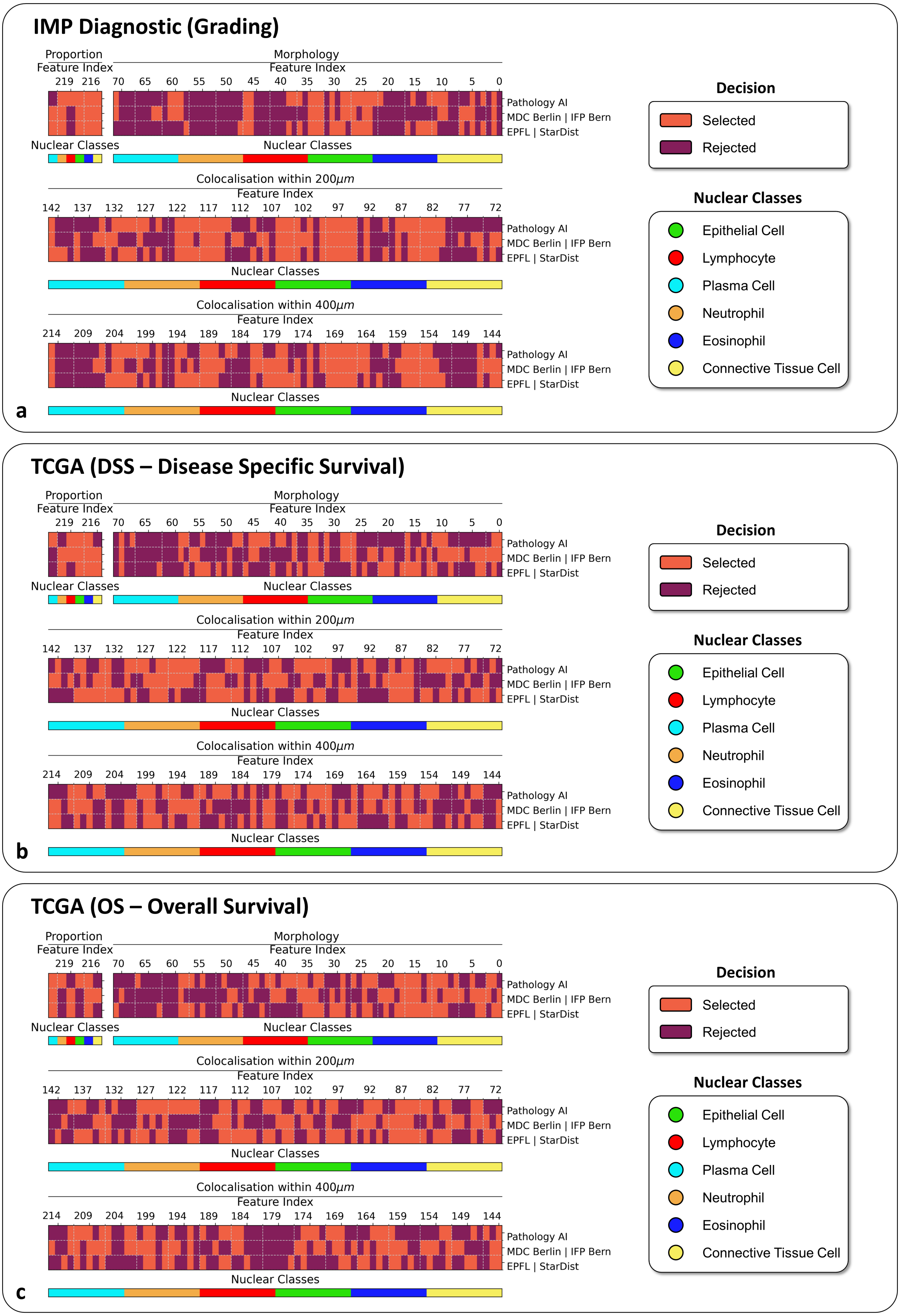

**Extended Data Fig. 5 | Summary on which features out of 222 extracted ones that were utilised for subsequent analyses on grading, disease specific survival or overall survival tasks. a**, **b** and **c** show the selected features for dysplasia grading, disease-specific survival analysis and overall survival analysis, respectively.

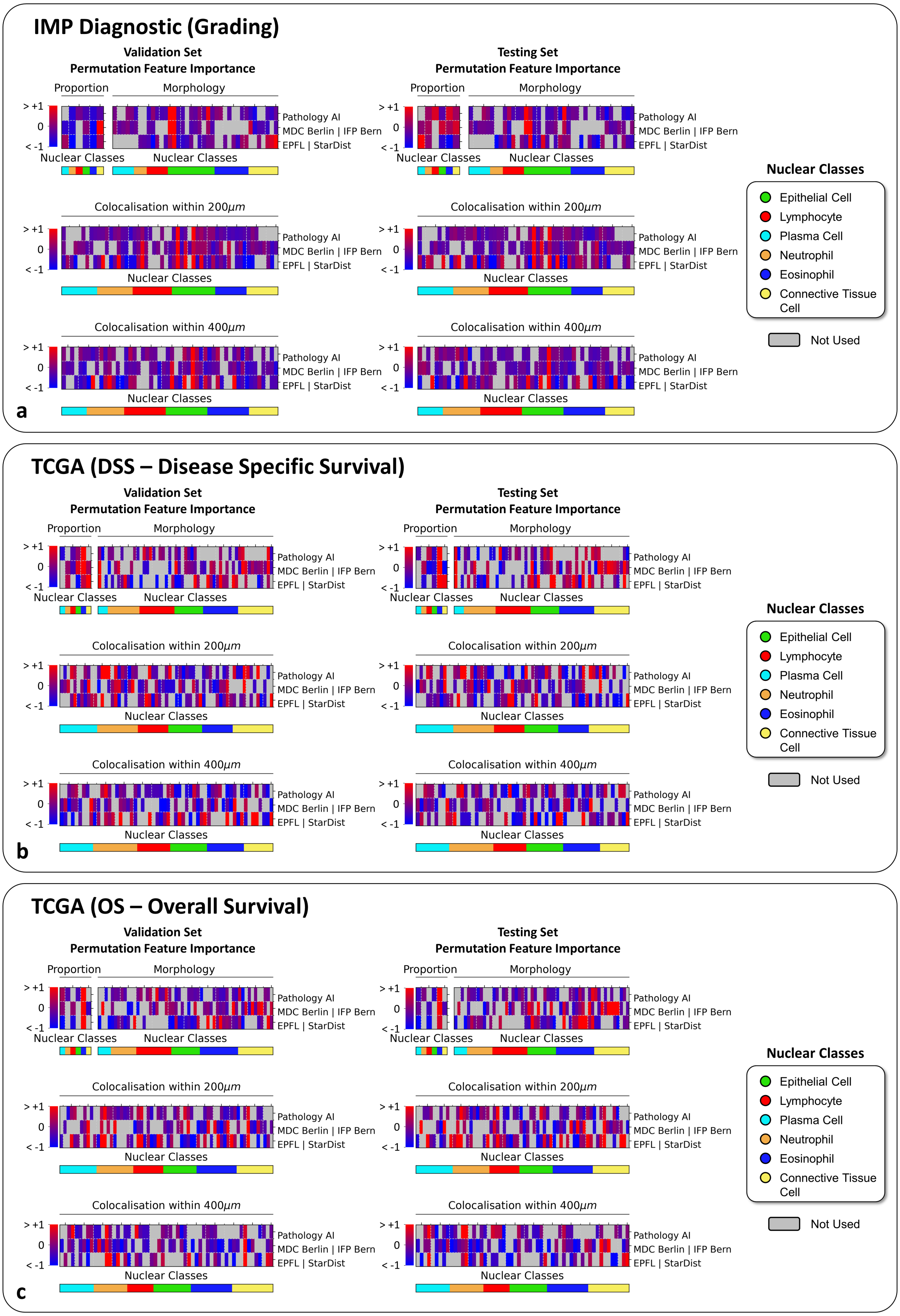

**Extended Data Fig. 6 | Summary on the overall importance of each selected feature in Extended Data Fig. 5 from each team for each task.** The importance is measured by a *permutation test* and combined across all data splits. A feature is important if the changes in its value heavily impact the *evaluation results* (C-Index for survival tasks and QWK for grading)**.** Features that are important to the final XGBoost model performance are coloured in red. On the other hand, a feature being blue indicates gains in performance when the its value become more noisy (undesirable or significantly less important). Grey cells show features not selected by this team but are selected by others.

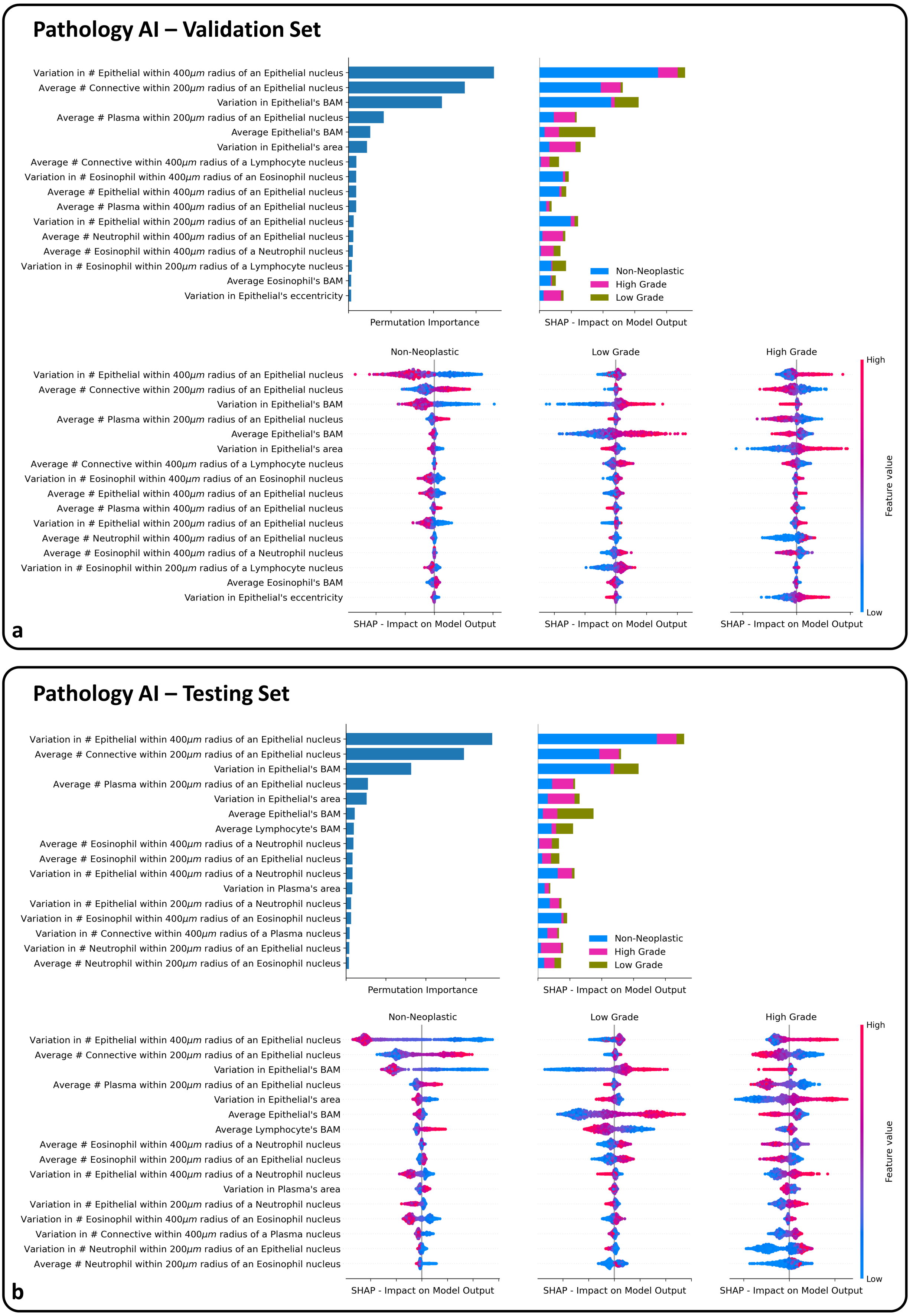

**Extended Data Fig. 7 | Contribution of the top 16 features from Pathology AI (taken from Extended Data Fig. 6) for the grading task on IMP Diagnostics dataset.** The first column reports the permutation importance of the feature on the *evaluation results* (QWK). On the other hand, other columns (SHAP) reflect how the changes in the feature value affect the *model predictions* (the predicted probabilities of each class by XGBoost). The reported importances are combined across all data splits.

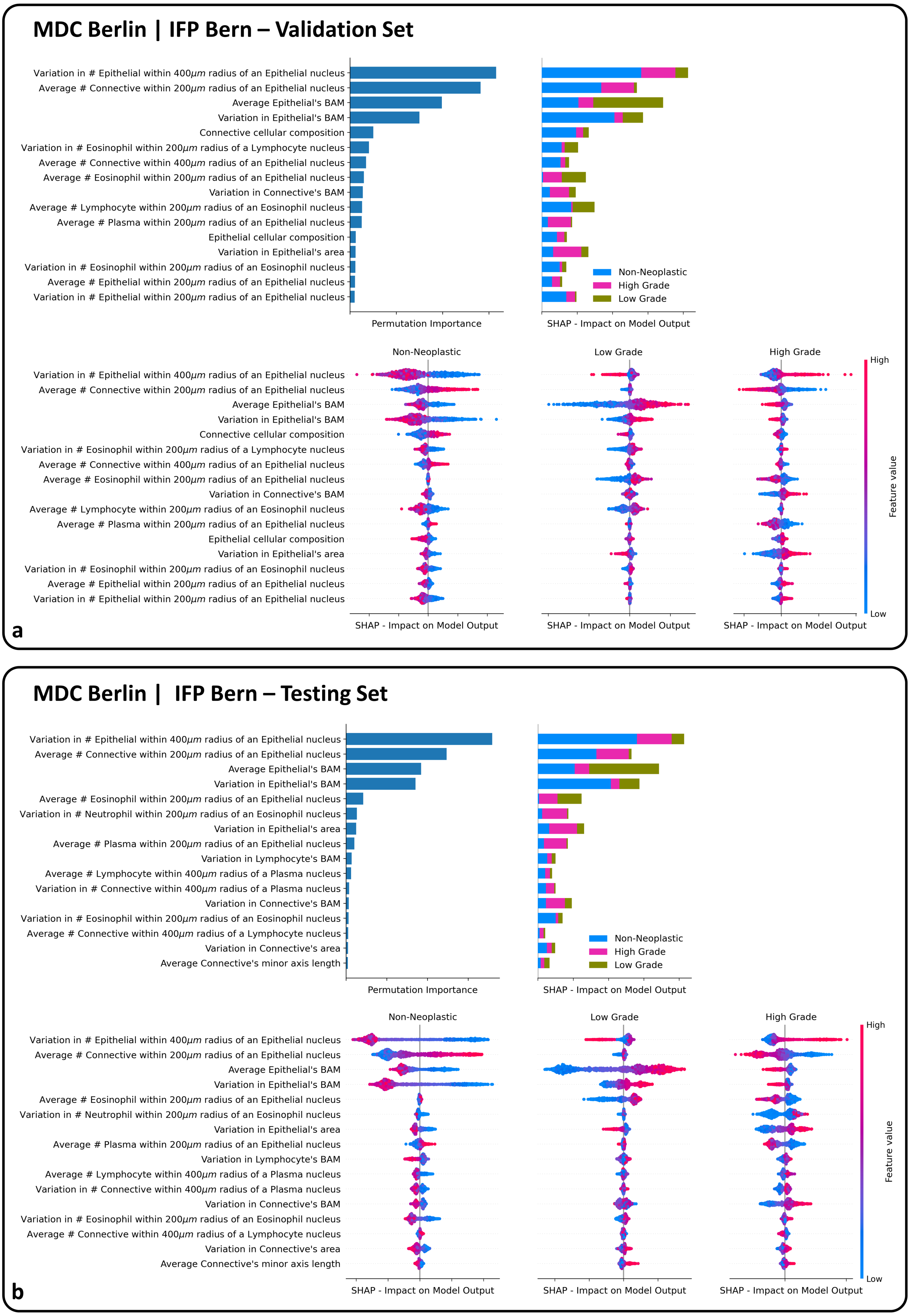

**Extended Data Fig. 8 | Contribution of the top 16 features from MDC Berlin | IFP Bern (taken from Extended Data Fig. 6) for the grading task on IMP Diagnostics dataset.** The first column reports the permutation importance of the feature on the *evaluation results* (QWK). On the other hand, other columns (SHAP) reflect how the changes in the feature value affect the *model predictions* (the predicted probabilities of each class by XGBoost). The reported importances are combined across all data splits.

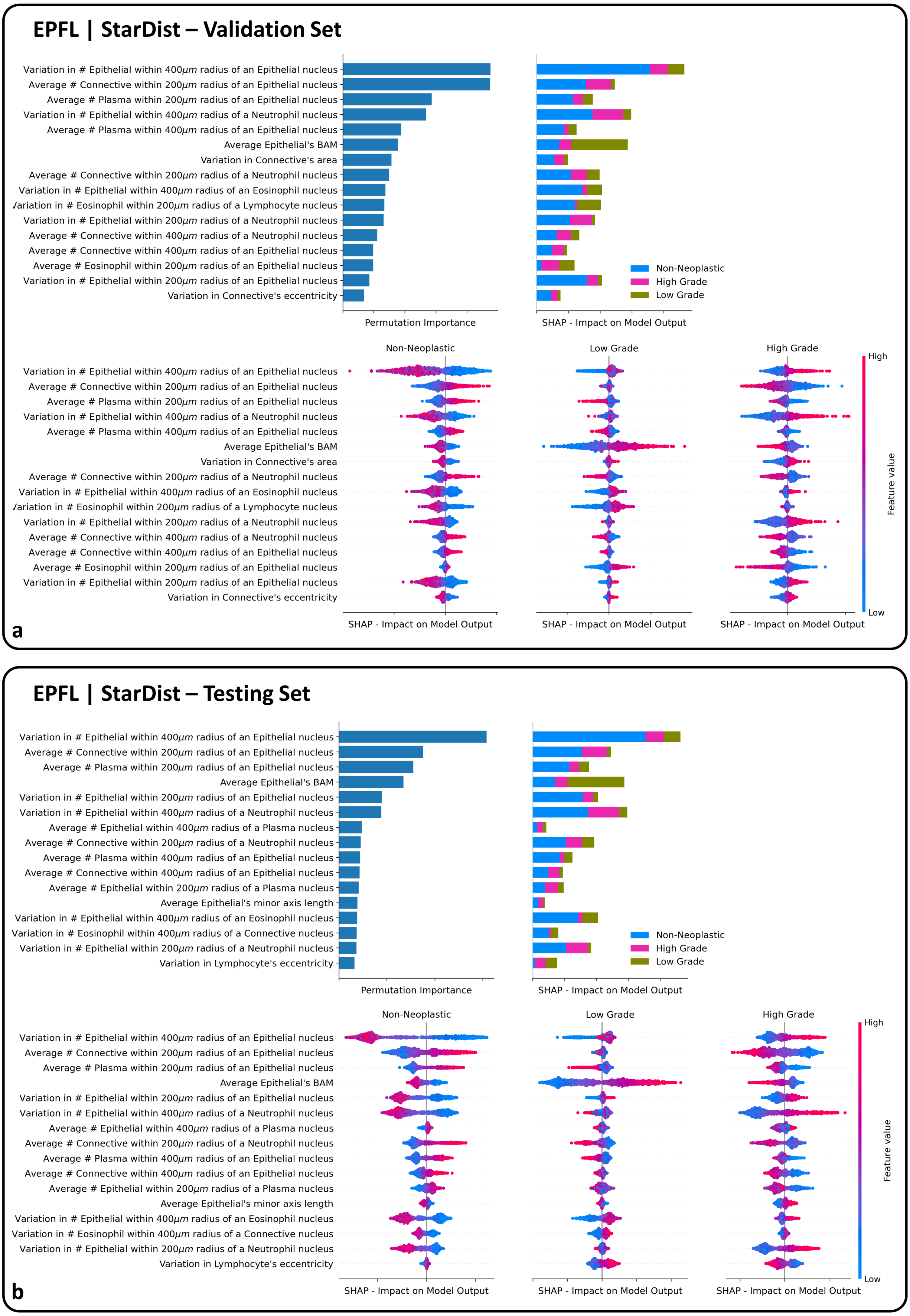

**Extended Data Fig. 9 | Contribution of the top 16 features from EPFL | StarDist (taken from Extended Data Fig. 6) for the grading task on IMP Diagnostics dataset.** The first column reports the permutation importance of the feature on the *evaluation results* (QWK). On the other hand, other columns (SHAP) reflect how the changes in the feature value affect the *model predictions* (the predicted probabilities of each class by XGBoost). The reported importances are combined across all data splits.

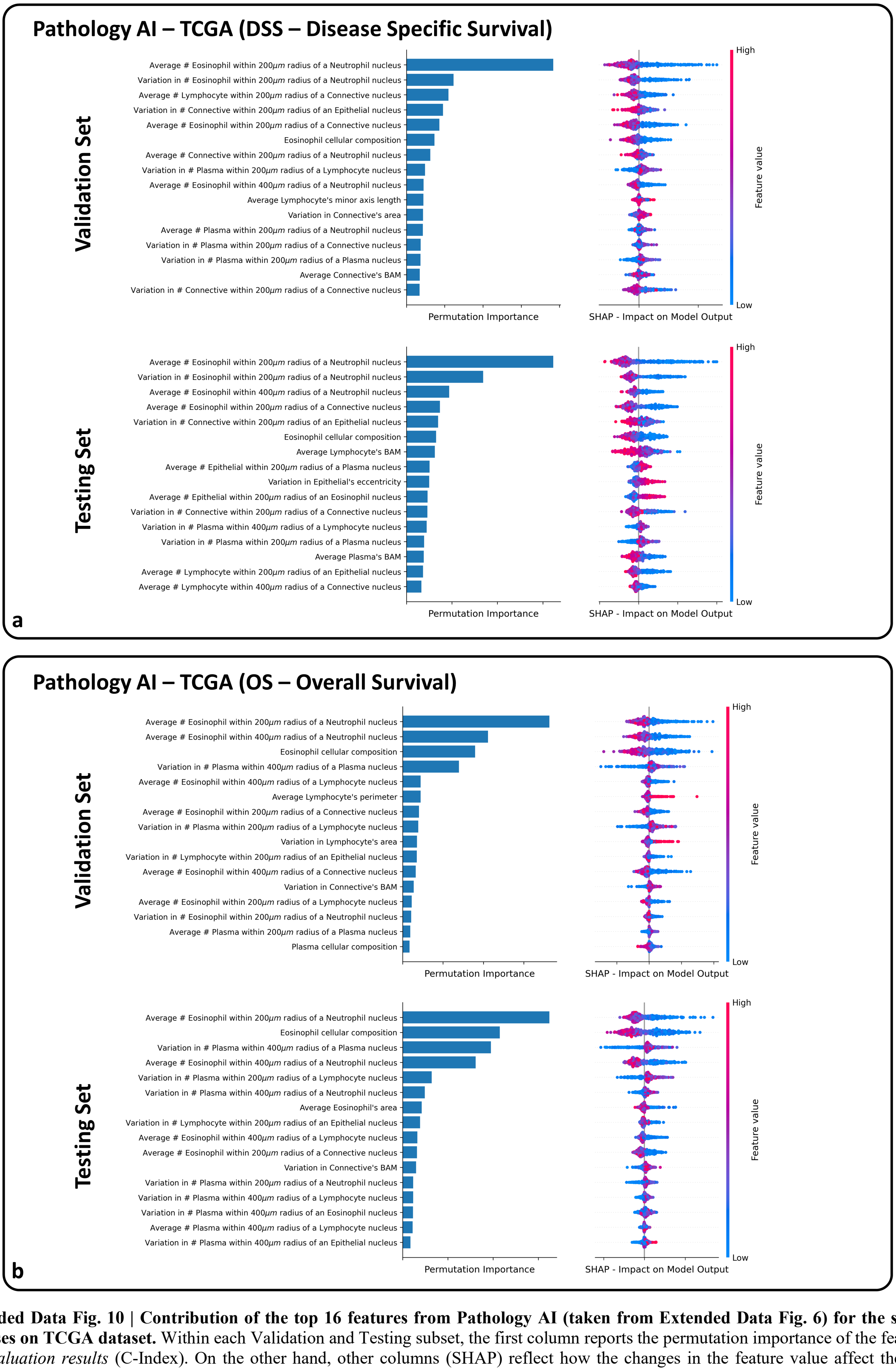

**Extended Data Fig. 10 | Contribution of the top 16 features from Pathology AI (taken from Extended Data Fig. 6) for the survival analyses on TCGA dataset.** Within each Validation and Testing subset, the first column reports the permutation importance of the feature on the *evaluation results* (C-Index). On the other hand, other columns (SHAP) reflect how the changes in the feature value affect the *model predictions* (the predicted risk scores by XGBoost). The reported importances are combined across all data splits.

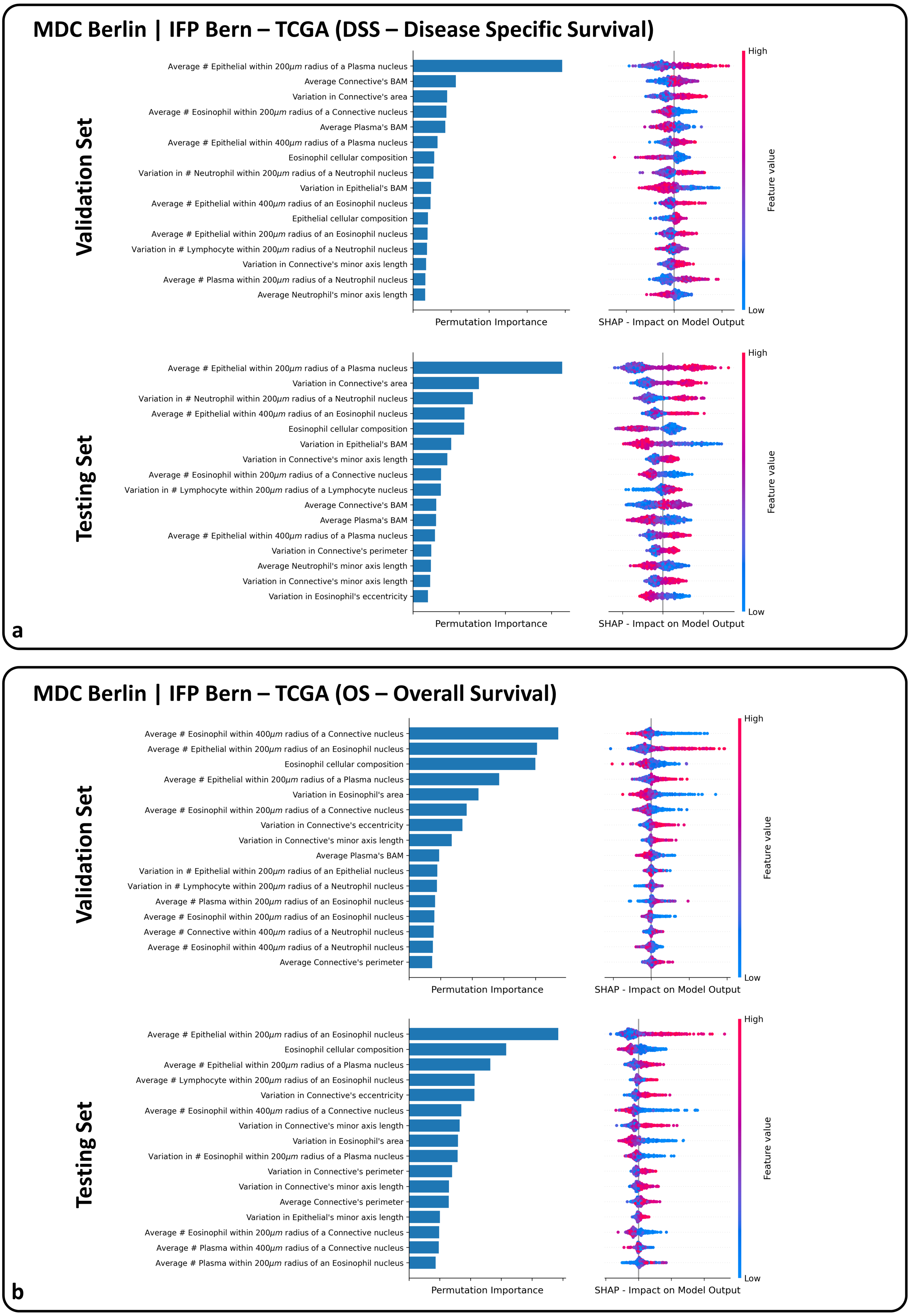

**Extended Data Fig. 11 | Contribution of the top 16 features from MDC Berlin | IFP Bern (taken from Extended Data Fig. 6) for the survival analyses on TCGA dataset.** Within each Validation and Testing subset, the first column reports the permutation importance of the feature on the *evaluation results* (C-Index). On the other hand, other columns (SHAP) reflect how the changes in the feature value affect the *model predictions* (the predicted risk scores by XGBoost). The reported importances are combined across all data splits.

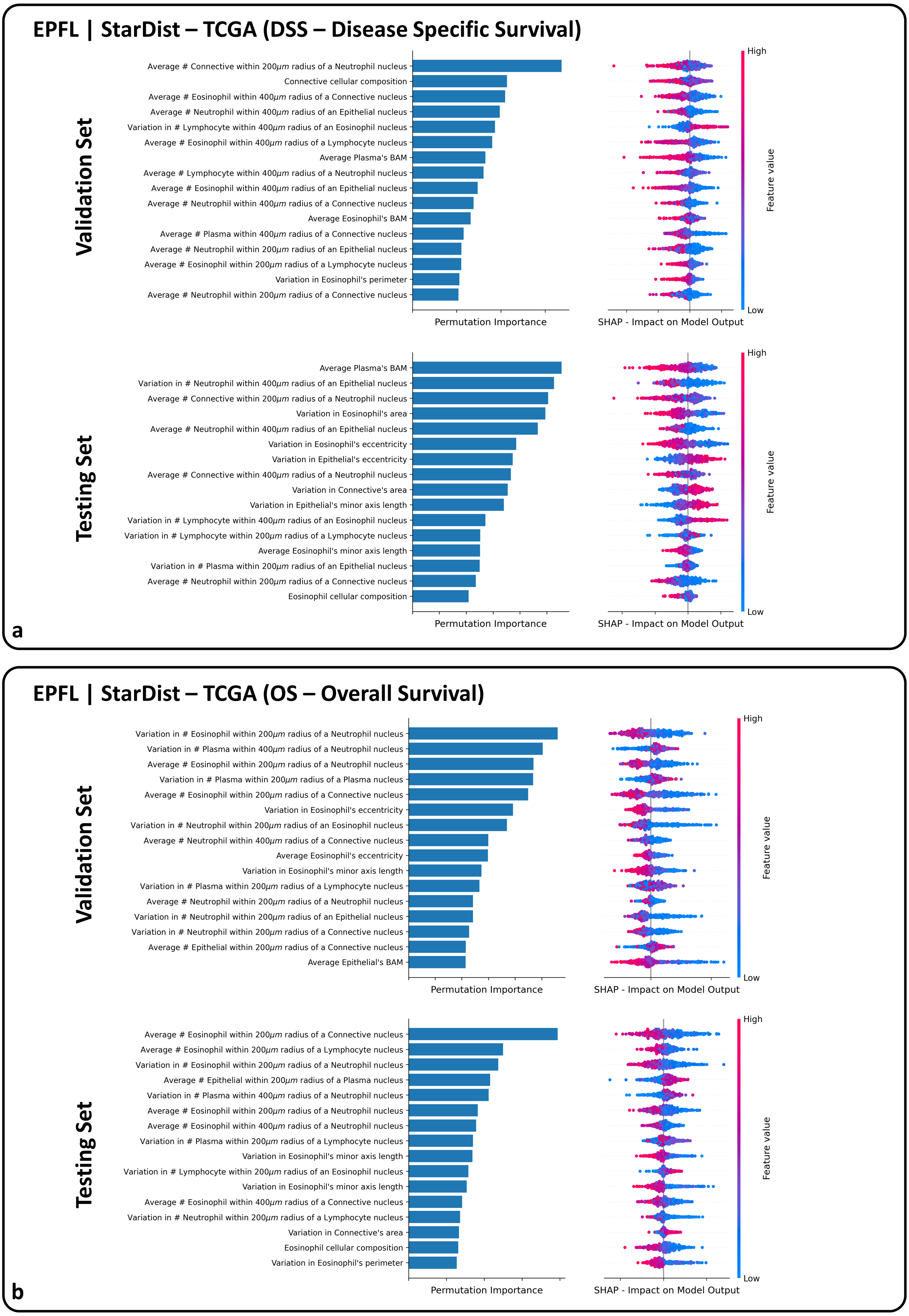

**Extended Data Fig. 12 | Contribution of the top 16 features from EPFL | StarDist (taken from Extended Data Fig. 6) for the survival analyses on TCGA dataset.** Within each Validation and Testing subset, the first column reports the permutation importance of the feature on the *evaluation results* (C-Index). On the other hand, other columns (SHAP) reflect how the changes in the feature value affect the *model predictions* (the predicted risk scores by XGBoost). The reported importances are combined across all data splits.

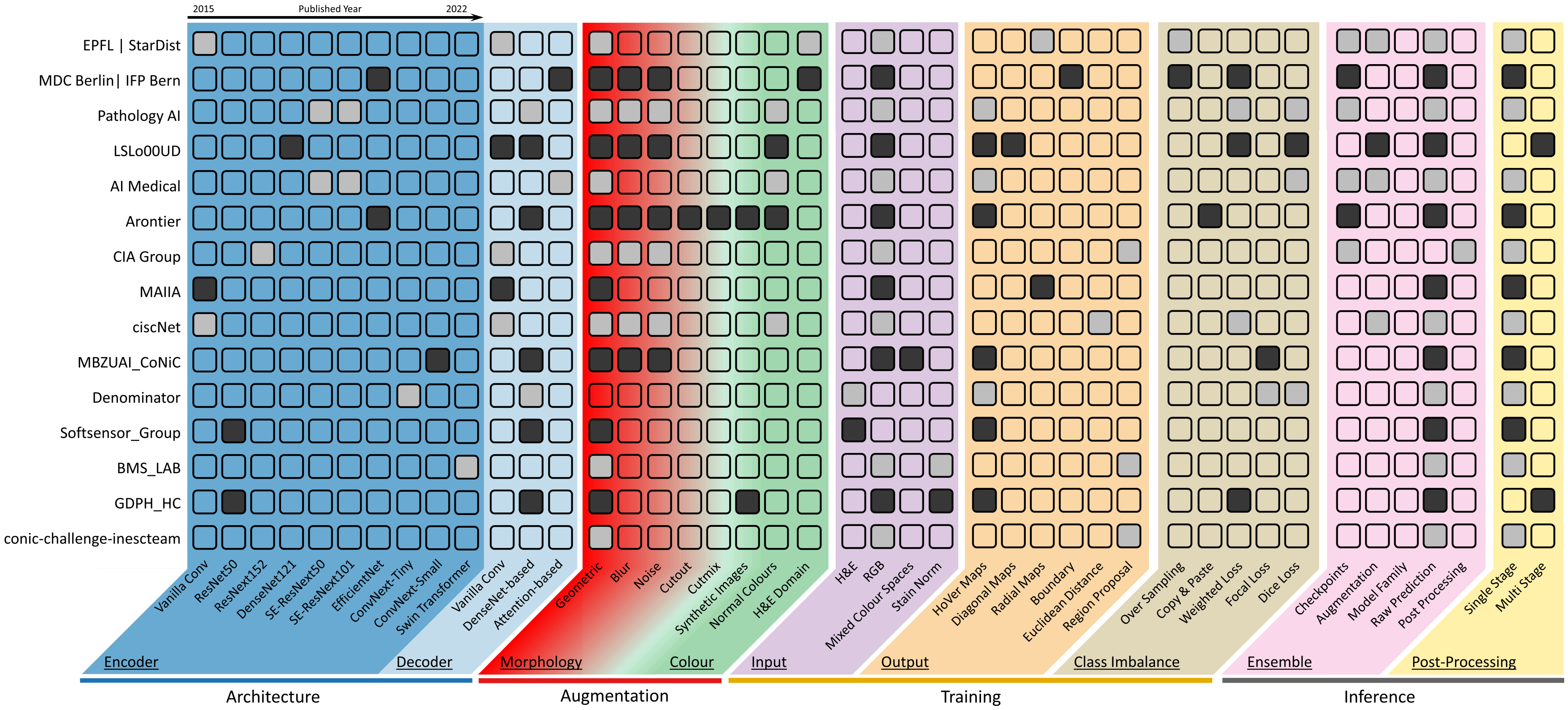

**Extended Data Fig. 13 | Summary of the top 15 participant algorithms.** The figure is split into various segments to better understand the difference between each team. Specifically, we identify the network architecture of each submission, including the encoder and decoder design. We determine the augmentation strategy, distinguishing between morphology-based and colour-based augmentation. We indicate the training strategy, consisting of the input type, the output type and whether a technique was used to deal with the class imbalance. We also identified the inference strategy, denoting whether ensembling was used and the post-processing technique. The colour within each box (grey or black) is insignificant – it is used to help distinguish between teams on each row.

# References

1. Rakha, E.A., *et al.* Prognostic significance of Nottingham histologic grade in invasive breast carcinoma. *Journal of clinical oncology* **26**, 3153-3158 (2008).
2. Lennard-Jones, J. Classification of inflammatory bowel disease. *Scandinavian Journal of Gastroenterology* **24**, 2-6 (1989).
3. Magro, F., *et al.* European consensus on the histopathology of inflammatory bowel disease. *Journal of Crohn's and Colitis* **7**, 827-851 (2013).
4. Ropponen, K.M., Eskelinen, M.J., Lipponen, P.K., Alhava, E. & Kosma, V.M. Prognostic value of tumour-infiltrating lymphocytes (TILs) in colorectal cancer. *The Journal of Pathology: A Journal of the Pathological Society of Great Britain and Ireland* **182**, 318-324 (1997).
5. Salgado, R., *et al.* The evaluation of tumor-infiltrating lymphocytes (TILs) in breast cancer: recommendations by an International TILs Working Group 2014. *Annals of oncology* **26**, 259-271 (2015).
6. Sahai, E., *et al.* A framework for advancing our understanding of cancer-associated fibroblasts. *Nature Reviews Cancer* **20**, 174-186 (2020).
7. Tommelein, J., *et al.* Cancer-associated fibroblasts connect metastasis-promoting communication in colorectal cancer. *Frontiers in oncology* **5**, 63 (2015).
8. Snead, D.R., *et al.* Validation of digital pathology imaging for primary histopathological diagnosis. *Histopathology* **68**, 1063-1072 (2016).
9. Graham, S., *et al.* Hover-net: Simultaneous segmentation and classification of nuclei in multi-tissue histology images. *Medical Image Analysis* **58**, 101563 (2019).
10. Schmidt, U., Weigert, M., Broaddus, C. & Myers, G. Cell detection with star-convex polygons. in *International Conference on Medical Image Computing and Computer-Assisted Intervention* 265-273 (Springer, 2018).
11. Gamper, J., Alemi Koohbanani, N., Benet, K., Khuram, A. & Rajpoot, N. Pannuke: an open pan-cancer histology dataset for nuclei instance segmentation and classification. in *European congress on digital pathology* 11-19 (Springer, 2019).
12. Gamper, J., *et al.* Pannuke dataset extension, insights and baselines. *arXiv preprint arXiv:2003.10778* (2020).
13. Koohbanani, N.A., Jahanifar, M., Tajadin, N.Z. & Rajpoot, N. NuClick: a deep learning framework for interactive segmentation of microscopic images. *Medical Image Analysis* **65**, 101771 (2020).
14. Graham, S., *et al.* Lizard: a large-scale dataset for colonic nuclear instance segmentation and classification. in *Proceedings of the IEEE/CVF International Conference on Computer Vision* 684-693 (2021).
15. Bejnordi, B.E., *et al.* Diagnostic assessment of deep learning algorithms for detection of lymph node metastases in women with breast cancer. *Jama* **318**, 2199-2210 (2017).
16. Bulten, W., *et al.* Automated deep-learning system for Gleason grading of prostate cancer using biopsies: a diagnostic study. *The Lancet Oncology* **21**, 233-241 (2020).
17. Sirinukunwattana, K., *et al.* Gland segmentation in colon histology images: The glas challenge contest. *Medical image analysis* **35**, 489-502 (2017).
18. Da, Q., *et al.* DigestPath: a Benchmark Dataset with Challenge Review for the Pathological Detection and Segmentation of Digestive-System. *Medical Image Analysis*, 102485 (2022).
19. Kumar, N., *et al.* A multi-organ nucleus segmentation challenge. *IEEE transactions on medical imaging* **39**, 1380-1391 (2019).
20. Vu, Q.D., *et al.* Methods for segmentation and classification of digital microscopy tissue images. *Frontiers in bioengineering and biotechnology*, 53 (2019).
21. Verma, R., *et al.* Multi-organ nuclei segmentation and classification challenge 2020. *IEEE transactions on medical imaging* **39**, 8 (2020).
22. Shaban, M., *et al.* Context-aware convolutional neural network for grading of colorectal cancer histology images. *IEEE transactions on medical imaging* **39**, 2395-2405 (2020).
23. Lu, M.Y., *et al.* AI-based pathology predicts origins for cancers of unknown primary. *Nature* **594**, 106-110 (2021).
24. Wulczyn, E., *et al.* Deep learning-based survival prediction for multiple cancer types using histopathology images. *PloS one* **15**, e0233678 (2020).

25. Zhu, W., Xie, L., Han, J. & Guo, X. The application of deep learning in cancer prognosis prediction. *Cancers* **12**, 603 (2020).
26. Kather, J.N., *et al.* Predicting survival from colorectal cancer histology slides using deep learning: A retrospective multicenter study. *PLoS medicine* **16**, e1002730 (2019).
27. Graham, S., *et al.* Conic: Colon nuclei identification and counting challenge 2022. *arXiv preprint arXiv:2111.14485* (2021).
28. Oliveira, S.P., *et al.* CAD systems for colorectal cancer from WSI are still not ready for clinical acceptance. *Scientific Reports* **11**, 1-15 (2021).
29. Deshpande, S., Dawood, M., Minhas, F. & Rajpoot, N. SynCLay: Interactive Synthesis of Histology Images from Bespoke Cellular Layouts. *arXiv preprint arXiv:2212.13780* (2022).
30. Hida, K., Maishi, N., Torii, C. & Hida, Y. Tumor angiogenesis—characteristics of tumor endothelial cells. *International journal of clinical oncology* **21**, 206-212 (2016).
31. Vu, Q.D., *et al.* Nuclear Segmentation and Classification: On Color and Compression Generalization. in *Machine Learning in Medical Imaging: 13th International Workshop, MLMI 2022, Held in Conjunction with MICCAI 2022, Singapore, September 18, 2022, Proceedings* 249-258 (Springer, 2022).
32. Foote, A., Asif, A., Rajpoot, N. & Minhas, F. REET: robustness evaluation and enhancement toolbox for computational pathology. *Bioinformatics* **38**, 3312-3314 (2022).
33. Shia, J., *et al.* Morphological characterization of colorectal cancers in The Cancer Genome Atlas reveals distinct morphology–molecular associations: clinical and biological implications. *Modern pathology* **30**, 599-609 (2017).
34. Schmitt, M. & Greten, F.R. The inflammatory pathogenesis of colorectal cancer. *Nature Reviews Immunology* **21**, 653-667 (2021).
35. Reichman, H., *et al.* Activated Eosinophils Exert Antitumorigenic Activities in Colorectal CancerEosinophils in Colorectal Cancer. *Cancer immunology research* **7**, 388-400 (2019).
36. Dawood, M., Branson, K., Rajpoot, N.M. & Minhas, F. Albrt: Cellular composition prediction in routine histology images. in *Proceedings of the IEEE/CVF International Conference on Computer Vision* 664-673 (2021).
37. Wahab, N., *et al.* Semantic annotation for computational pathology: Multidisciplinary experience and best practice recommendations. *The Journal of Pathology: Clinical Research* **8**, 116-128 (2022).
38. Awan, R., *et al.* Glandular morphometrics for objective grading of colorectal adenocarcinoma histology images. *Scientific reports* **7**, 1-12 (2017).
39. Berry, S., *et al.* Analysis of multispectral imaging with the AstroPath platform informs efficacy of PD-1 blockade. *Science* **372**, eaba2609 (2021).
40. Vu, Q.D., Rajpoot, K., Raza, S.E.A. & Rajpoot, N. Handcrafted Histological Transformer (H2T): Unsupervised Representation of Whole Slide Images. *arXiv preprint arXiv:2202.07001* (2022).
41. Chen, T. & Guestrin, C. Xgboost: A scalable tree boosting system. in *Proceedings of the 22nd acm sigkdd international conference on knowledge discovery and data mining* 785-794 (2016).
42. Kira, K. & Rendell, L.A. A practical approach to feature selection. in *Machine learning proceedings 1992* 249-256 (Elsevier, 1992).
43. Kursa, M.B. & Rudnicki, W.R. Feature selection with the Boruta package. *Journal of statistical software* **36**, 1-13 (2010).
44. Altmann, A., Toloşi, L., Sander, O. & Lengauer, T. Permutation importance: a corrected feature importance measure. *Bioinformatics* **26**, 1340-1347 (2010).

# Supplementary Material

## S1. Preliminary test phase results

As well as preparing participants for their final submissions, the preliminary submission phase provided an opportunity to assess and improve the performance of algorithms developed during the discovery phase of the competition. The segmentation and classification task was ranked by the multi-class panoptic quality ($mPQ^+$)[27], which takes the average of the per-class panoptic quality scores. The cellular composition task was ranked by the multi-class coefficient of determination ($mR^2$), which similarly takes the average of the per-class scores. In Extended Data Fig. 1 we show the results over the course of the preliminary submission phase for both tasks. In parts *a* and *c* of the figure, we show the average results of all submissions per day over time, along with the corresponding error bars. In parts *b* and *d,* we show the best performance per day over time. Generally, we can see that the preliminary submission phase was successful in helping to improve the performance of submitted models, due to its competitive nature.

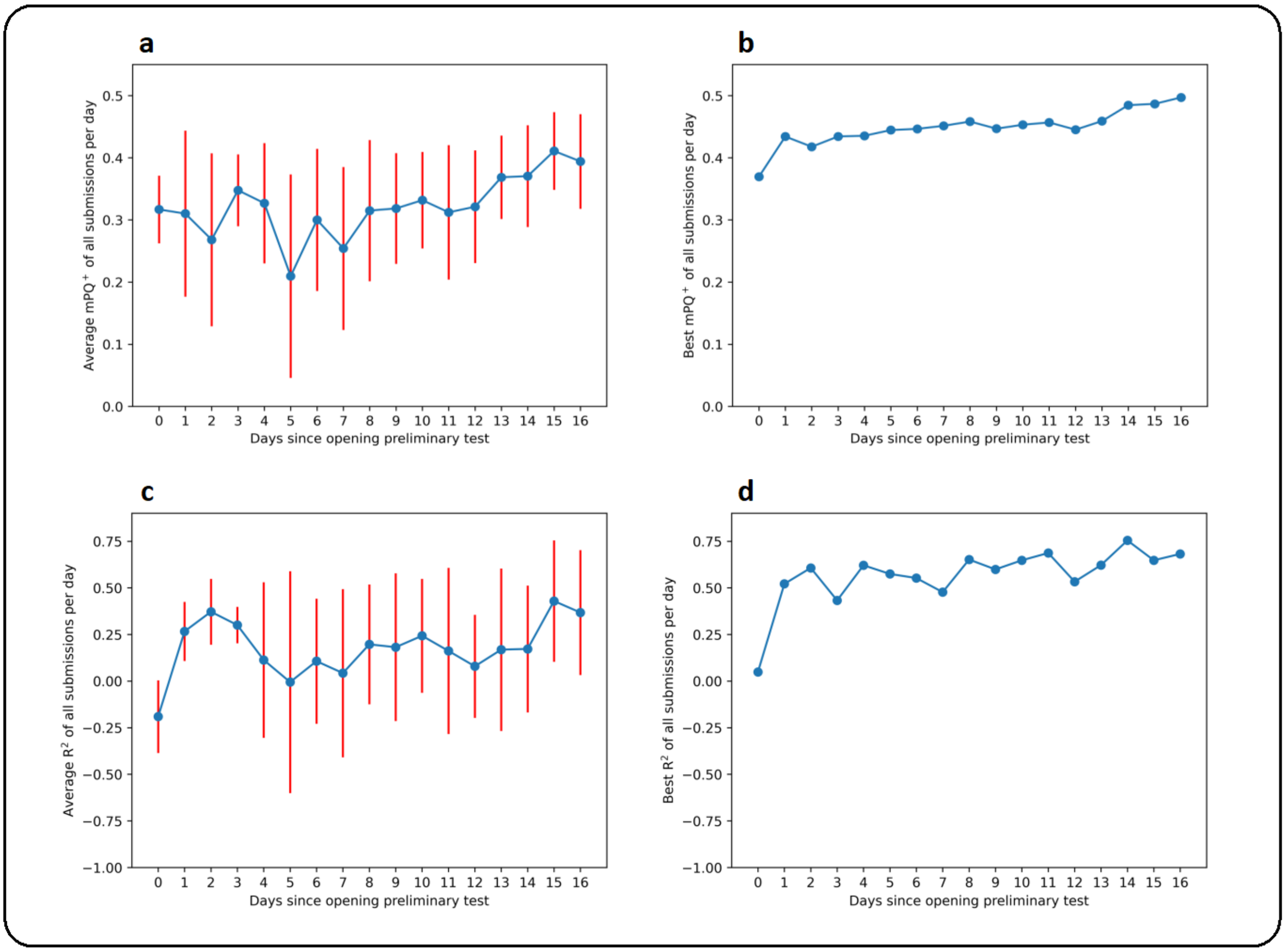

**Supplementary Fig. 1 | Results during the preliminary submission phase of the competition for both tasks. a**, Average score per day for the segmentation task; **b**, best score per day for the segmentation task; **c**, average score per day for the cellular composition task; **d**, best score per day for the cellular composition task.

## S2. Detailed summary of the challenge algorithms

We received 26 and 24 submissions to the final segmentation and classification and cellular composition leaderboards, respectively. At the time of submission, we required all participants to submit a short paper outlining their approach. These can be viewed, by visiting the final test leaderboards at the following web page: https://warwick.ac.uk/conic-challenge. While each technique is described in detail in the provided manuscripts, we also outline a summary of the submitted methods below. Here, we give an overview of the model architecture, the instance segmentation target, the loss function and whether a strategy was used to overcome the class imbalance present in the dataset. In the descriptions below, SC denotes segmentation and classification, while CC denotes cellular composition.

### EPFL | StarDist

**Leaderboard: SC = 1st, CC = 3rd**

StarDist[1] is based on U-Net[2] architecture and it predicts an object probability map and 64 radial distance maps. The conventional StarDist model does not perform nuclear classification – therefore for this challenge, a second upsampling branch was added to perform semantic segmentation. To deal with the class imbalance, patches that contained minority classes were oversampled during training. Geometric and H&E-based augmentations were used and multiple models were ensembled with test-time augmentation to obtain the final prediction.

### MDC Berlin | IFP Bern

**Leaderboard: SC = 2nd, CC = 9th**

A U-Net style architecture was used with an EfficientNet[3] backbone and two upsampling branches. The first branch performed instance segmentation and the second branch performed semantic segmentation. The instance segmentation branch predicted each pixel to be either: the interior of the nucleus; the nuclear boundary; or the background. In addition, regression of the nuclear centroids was performed as an auxiliary task. For class imbalance, oversampling of patches containing underrepresented nuclear classes was performed in combination with utilisation of a weighted focal loss. Geometric, blur, noise and H&E-based augmentations were used and multiple models were ensembled.

### Pathology AI

**Leaderboard: SC = 3rd, CC = 1st**

A HoVer-Net[4] was used with a SE-ResNeXt101[5] backbone and heavy dropout[6] layers in the upsampling branches. Despite describing the concept of diagonal distance maps in their method description paper, standard horizontal and vertical distance maps[4] were used in the final submission. A combination of dice and weighted cross entropy loss was used to help overcome the class imbalance. Geometric, blur and colour augmentations were used during training. The final model was trained on several splits of the data and the results ensembled for submission.

### LSLL000UD

**Leaderboard: SC = 4th, CC = 6th**

A HoVer-Net[4] with a DenseNet-121[7] backbone was used for instance segmentation, but without the classification branch for simultaneous prediction. Diagonal distance maps were used to improve the performance. A second lightweight U-Net[2] was used to perform boundary refinement, which takes probability maps cropped at each nucleus as input. Then, a devoted network for pixel-wise nuclear classification was used with the same base architecture as the HoVer-Net for instance segmentation. A combination of standard cross entropy and dice loss were used to deal with the class imbalance. Geometric, blur, noise and colour augmentations were performed during training and test-time augmentations were used to obtain the final result.

### AI_medical

**Leaderboard: SC = 5th, CC = 2nd**

A HoVer-Net[4] was utilised with an SE-ResNeXt50[5] backbone and a Coordinate Attention module[8] in the decoder. Conventional horizontal and vertical distance maps were used to perform instance segmentation. To counter the class imbalance, the submission utilised a both dice and weighted cross entropy loss in the classification branch. Geometric and colour augmentations were used during optimisation. For the final submission several models were ensembled and test-time augmentation were used.

## Arontier

**Leaderboard: SC = 6th, CC = 7th**

A HoVer-Net[4] model was used, with skip connections inspired by U-Net++[9] and an EfficientNet[3] backbone. An interesting approach was used for dealing with the class imbalance, consisting of copy and paste augmentation[10]. Geometric, blur, noise, Cutout[11], Cutmix[12] and colour augmentations were used and an ensemble of 5 models trained on different data splits was considered for the final submission.

## CIA Group

**Leaderboard: SC = 7th, CC = 4th**

An ensemble of a conventional HoVer-Net[4] and Cascaded Mask-RCNN[13], both with ResNeXt-152[14] backbones, was used for the challenge. However, despite this strategy being used during the preliminary submission phase, it exceeded the maximum 60-minute allotted time during the final submission phase. Therefore, the final submission comprised of the Cascaded Mask-RCNN by itself. No method was documented for dealing with the class imbalance in the dataset. Geometric, blur and noise augmentations were used during training and model ensembling was performed when making the submission.

## MAIIA

**Leaderboard: SC = 8th, CC = 17th**

A StarDist[1] model with a U-Net[2] architecture was implemented but using more convolutional filters than the standard approach. Here, the off-the-shelf StarDist repository was used to see how it performed with minimal modification. The model predicted the star convex polygons for each nucleus by outputting an object probability map, along with 32 radial distance maps. Fairly conventional augmentations, consisting of geometric transformations and additive noise were used. No specific strategy was utilised for dealing with the class imbalance and no form of ensembling was performed.

## ciscNet

**Leaderboard: SC = 9th, CC = 11th**

A regular U-Net[2] architecture was used, but with group normalisation[15] and mish activation[16]. The normalised Euclidean distance maps of nuclear pixels to their nearest boundary was predicted to enable instance segmentation. For this, a separate distance map was considered per nuclear type to enable simultaneous classification. To counter the class imbalance, a weighted summation of the per-class regression losses was utilised, where more weight was given to minority classes. Geometric, blur, noise and colour augmentations were used during training and test-time augmentation was performed to yield the final result.

## MBZUAI_CoNIC

**Leaderboard: SC = 10th, CC = 8th**

A HoVer-Net[4] with a ConvNeXt-Small backbone[17] was used with standard horizontal and vertical distance maps as the instance segmentation target. With an aim of learning more discriminative features, each image was converted to various colour spaces and concatenated with the original RGB image before input to the network. A unified focal loss[18] was used during training, which aimed to counter the class imbalance. Geometric, noise and blur augmentations were used during training.

## Denominator

**Leaderboard: SC = 11th, CC = 10th**

Similar to above, HoVer-Net[4] with a ConvNeXt-Tiny backbone[17] was used. Separation of the Haematoxylin and eosin stains was performed before input to the network. A combination of focal[19] and dice loss was used to help combat the imbalance of classes in the data. No model ensembling was used during submission of the algorithm.

## Softsensor_Group

**Leaderboard: SC = 12th, CC = 5th**

A fusion of HoVer-Net[4] and Triple U-Net[20] was used that considered both the original RGB image and the Haematoxylin stain channel as input. Each input was processed by a separate encoder, which were then fused using a progressive dense feature aggregation block. Following HoVer-Net, the model predicted the horizontal and vertical maps, binary nuclear segmentation map, and the multi-class semantic segmentation map. All RGB input patches used Reinhard normalisation[21] to combat differences in the stain appearance and geometric augmentations were introduced during training.

## BMS_LAB

**Leaderboard: SC = 13th, CC = 12th**

A Swin-Transformer[22] with a Hybrid Task Cascade model[23] was used. The model did not use a strategy to deal with the class imbalance. Geometric augmentations were performed and Macenko stain normalisation[24] was used to help reduce the variability of the image appearance across the dataset. For submission, input images were resized to five different scales before processing and the results were then merged together.

## GDPH_HC

**Leaderboard: SC = 14th, CC = 13th**

HoVer-Net[4] was used without any modification to the original architecture. To enhance the available data for training, a generative adversarial network[25] was used to create synthetic images as an augmentation strategy. In addition, a self-supervised technique called RestainNet[26] was used to perform stain normalisation and geometric transformations were applied to all input images. To help train with the presence imbalanced data, a class-weighted loss function was incorporated at the output of the classification branch.

## conic-challenge-inescteam

**Leaderboard: SC = 15th**

CenterNet[27] was used, which is a probabilistic two-stage object detection model. This model allows the reduction of proposals from the Region Proposal Network (RPN), which could be important in this application where each image has many objects. Like Mask-RCNN[28], the original CenterNet approach waas extended so that is also produced a segmentation mask for each nucleus. No specific method was used to deal with class imbalance and geometric augmentation was used during model training.

## Aman

**Leaderboard: SC = 16th**

A subtly modified HoVer-Net[4] model was used for the challenge, where major focus given to the data preprocessing step. Copy and paste augmentation[10] of neutrophil and eosinophil nuclei was utilised in addition to performing geometric augmentation of the images. Also, a transformation of the colour space of images was applied to increase the variability of the stain appearance in the training set. Following this, weighted cross entropy and weighted dice loss functions were used to help counter the class imbalance in the dataset.

## Bin

**Leaderboard: SC = 17th, CC = 19th**

A HoVer-Net[4] approach was used, but each convolution was swapped with a multiple filter block. Here, multiple filter sizes were utilised in parallel during each operation and the results were merged. This was repeated throughout the network. A combination of cross entropy and Dice loss were used, like in the original HoVer-Net implementation and no augmentation was performed.

## DH-Goods

**Leaderboard: SC = 18th, CC = 16th**

Two separate HoVer-Nets[4] with the same architecture were used that aimed to tackle the class imbalance present in the dataset – one that considered epithelial, lymphocyte and connective tissue cell classes, and the other that considered plasma cell, neutrophil and eosinophil classes. The intuition was that separating out the minority classes may lead to better performance. Each HoVer-Net used a HR-Net backbone[29] with an Atrous Spatial Pyramid Pooling (ASPP) unit[30] after the encoder. In addition, a YOLOv5[31] was trained for nuclear detection and classification, where a U-Net model was used to generate the segmentation masks within the bounding boxes. For tackling the class imbalance, equalised focal loss was used during optimisation of YOLOv5 and mosaic augmentation was used as a way of introducing underrepresented classes into input images. HoVer-Net and YOLOv5 results were then merged using a custom strategy. Geometric transformations of input images were performed and test-time augmentation used to generate the final submission.

## VNIT

**Leaderboard: SC = 19th, CC = 22nd**

A hybrid approach was implemented incorporating handcrafted features, such as local binary patterns and histogram of oriented gradients, into a HoVer-Net[4] model. Specifically, handcrafted features were combined with the deep features after passing input images through the encoder and are then upsampled via three separate upsampling branches, in the same way as the original HoVer-Net approach. The same loss strategy as the original implementation was used and so no proposed technique was used to deal with the class imbalance. Blur augmentation and colour jitter was used during training.

## Sk

**Leaderboard: SC = 20th, CC = 8th**

An Eff-UNet[32] was used, which combines the effectiveness of EfficientNet[3] as the encoder with U-Net[2] as the decoder. The approach outputs two prediction maps: 1) direction map for instance segmentation and 2) semantic segmentation map for classification. The direction map divides each nucleus into $N$ segments around the centroid. For this submission, $N$ was set to be 4 and therefore divided each nucleus into quadrants. Each quadrant was then treated as a separate class to predict and instance segmentation was performed using a purpose-built post-processing pipeline. Colour, geometric and blur augmentation was used during training. A combination of cross entropy and dice loss was used for the semantic segmentation map, which may partly help alleviate the difficulty in dealing with the class imbalance. Rather than using the segmentation and classification output to predict cellular composition, a separate branch was added to the encoder that directly regressed the nuclear counts.

## Jiffy Labs and CET CV Lab

**Leaderboard: SC = 22nd, CC = 15th**

A HoVer-Net[4] with a ConvNeXt backbone[17] was used for the challenge submission. To help deal with the class imbalance, a combination of Dice and focal loss[19] was used.

## TIA Warwick

**Leaderboard: SC = 23rd, CC = 14th**

ALBRT[33] with an Xception backbone[34] was used for directly predicting the cellular composition from the input image, without performing nuclear segmentation. As opposed to the original approach that used a ranking loss, a Huber loss was used, that aimed to directly maximise the $R^2$ score. Due to the difficulty in predicting underrepresented classes, a separate network was trained for predicting the counts of eosinophils. Then, each network was trained multiple times and the per-class nuclear counts averaged for the final submission. A standard HoVer-Net[4] model was trained for the segmentation and classification task on a single split of the data.

## QuIIL

### Leaderboard: CC = 23rd

A YOLOv5[31] with a Cross Stage Partial Network backbone[35] was used to perform the task of nuclear detection and classification, where the results were then utilised to perform cellular composition. Geometric, colour and mosaic augmentations were used during training.

# S3. Downstream results

**Supplementary Table 1 | Performance of the XGBoost on TCGA dataset for DSS (disease specific survival) task when using feature set obtained from different nuclear recognition methods.** The reported values are mean ± std of C-index. $C$ denotes, clinical features, $D_d$ denotes density-based features, $D_m$ denotes morphology-based features and $D_c$ denotes colocalisation features. $D$ refers to the combination of all types of features (excluding clinical) and $D$ is a subset of $\bar{D}$ after feature selection.

**Validation Set**

| | $C$ | $D_d$ | $D_m$ | $D_c$ | $D$ | $\bar{D}$ |
|---|---|---|---|---|---|---|
| **N/A** | 0.7665±0.0600 | - | - | - | - | - |
| **Baseline** | - | 0.6270±0.0631 | 0.6018±0.0511 | 0.5922±0.0690 | 0.5981±0.0464 | 0.6228±0.0601 |
| **Pathology AI** | - | 0.6437±0.0648 | 0.5879±0.0683 | 0.6443±0.0492 | 0.6366±0.0650 | 0.6672±0.0583 |
| **MDC Berlin \| IFP Bern** | - | 0.6333±0.0482 | 0.6242±0.0732 | 0.6346±0.0591 | 0.6413±0.0604 | 0.6686±0.0443 |
| **EPFL \| StarDist** | - | 0.6418±0.0543 | 0.6087±0.0676 | 0.6334±0.0526 | 0.6354±0.0561 | 0.6685±0.0628 |

**Testing Set**

| | $C$ | $D_d$ | $D_m$ | $D_c$ | $D$ | $\bar{D}$ |
|---|---|---|---|---|---|---|
| **N/A** | 0.7662±0.0527 | - | - | - | - | - |
| **Baseline** | - | 0.6320±0.0772 | 0.5785±0.0747 | 0.5537±0.0782 | 0.5781±0.0634 | 0.5744±0.0738 |
| **Pathology AI** | - | 0.6284±0.0734 | 0.5742±0.0697 | 0.6263±0.0781 | 0.6106±0.0671 | 0.6450±0.0703 |
| **MDC Berlin \| IFP Bern** | - | 0.6081±0.0792 | 0.6358±0.0591 | 0.6088±0.0566 | 0.6160±0.0571 | 0.6518±0.0582 |
| **EPFL \| StarDist** | - | 0.6068±0.0826 | 0.6144±0.0722 | 0.5976±0.0978 | 0.6114±0.0827 | 0.6554±0.0631 |

**Supplementary Table 2 | Performance of the XGBoost on TCGA dataset for OS (overall survival) task when using feature set obtained from different nuclear recognition methods.** The reported values are mean ± std of C-index. $C$ denotes, clinical features, $D_d$ denotes density-based features, $D_m$ denotes morphology-based features and $D_c$ denotes colocalisation features. $D$ refers to the combination of all types of features (excluding clinical) and $D$ is a subset of $\bar{D}$ after feature selection.

**Validation Set**

| | $C$ | $D_d$ | $D_m$ | $D_c$ | $D$ | $\bar{D}$ |
|---|---|---|---|---|---|---|
| **N/A** | 0.7354±0.0481 | - | - | - | - | - |
| **Baseline** | - | 0.5888±0.0631 | 0.5478±0.0674 | 0.5769±0.0768 | 0.5748±0.0348 | 0.6015±0.0580 |
| **Pathology AI** | - | 0.6291±0.0543 | 0.5855±0.0623 | 0.6340±0.0558 | 0.6399±0.0426 | 0.6716±0.0526 |
| **MDC Berlin \| IFP Bern** | - | 0.6160±0.0476 | 0.6049±0.0727 | 0.6014±0.0613 | 0.6156±0.0644 | 0.6453±0.0533 |
| **EPFL \| StarDist** | - | 0.6140±0.0685 | 0.6051±0.0778 | 0.6275±0.0514 | 0.6349±0.0617 | 0.6589±0.0563 |

**Testing Set**

| | $C$ | $D_d$ | $D_m$ | $D_c$ | $D$ | $\bar{D}$ |
|---|---|---|---|---|---|---|
| **N/A** | 0.7251±0.0411 | - | - | - | - | - |
| **Baseline** | - | 0.5641±0.0731 | 0.5401±0.0757 | 0.5550±0.0649 | 0.5586±0.0647 | 0.5729±0.0650 |
| **Pathology AI** | - | 0.6185±0.0685 | 0.5838±0.0653 | 0.6218±0.0620 | 0.6187±0.0639 | 0.6418±0.0565 |
| **MDC Berlin \| IFP Bern** | - | 0.6019±0.0640 | 0.5808±0.0458 | 0.5876±0.0587 | 0.6115±0.0797 | 0.6176±0.0616 |
| **EPFL \| StarDist** | - | 0.5811±0.0719 | 0.5930±0.0688 | 0.5846±0.0755 | 0.6105±0.0653 | 0.6456±0.0614 |

**Supplementary Table 3 | Performance of the XGBoost on IMP Diagnostic for grading task when using feature set obtained from different nuclear recognition methods.** The reported values are mean ± std of $mF_1$. $C$ denotes, clinical features, $D_d$ denotes density-based features, $D_m$ denotes morphology-based features and $D_c$ denotes colocalisation features. $D$ refers to the combination of all types of features (excluding clinical) and $D$ is a subset of $\overline{D}$ after feature selection.

**Validation Set**

| | $\boldsymbol{D_d}$ | $\boldsymbol{D_m}$ | $\boldsymbol{D_c}$ | $\boldsymbol{D}$ | $\boldsymbol{\overline{D}}$ |
|---|---|---|---|---|---|
| **Baseline** | 0.6958±0.0212 | 0.7564±0.0265 | 0.7781±0.0230 | 0.8228±0.0185 | 0.8307±0.0208 |
| **Pathology AI** | 0.7565±0.0246 | 0.7702±0.0318 | 0.8520±0.0204 | 0.8669±0.0168 | 0.8720±0.0130 |
| **MDC Berlin \| IFP Bern** | 0.7110±0.0265 | 0.7442±0.0344 | 0.8545±0.0242 | 0.8705±0.0220 | 0.8765±0.0214 |
| **EPFL \| StarDist** | 0.7716±0.0234 | 0.7390±0.0265 | 0.8461±0.0169 | 0.8533±0.0186 | 0.8573±0.0184 |

**Testing Set**

| | $\boldsymbol{D_d}$ | $\boldsymbol{D_m}$ | $\boldsymbol{D_c}$ | $\boldsymbol{D}$ | $\boldsymbol{\overline{D}}$ |
|---|---|---|---|---|---|
| **Baseline** | 0.6862±0.0379 | 0.7402±0.0287 | 0.7729±0.0255 | 0.8203±0.0316 | 0.8227±0.0273 |
| **Pathology AI** | 0.7492±0.0349 | 0.7618±0.0317 | 0.8451±0.0296 | 0.8649±0.0282 | 0.8664±0.0280 |
| **MDC Berlin \| IFP Bern** | 0.6885±0.0297 | 0.7472±0.0304 | 0.8520±0.0264 | 0.8698±0.0256 | 0.8739±0.0218 |
| **EPFL \| StarDist** | 0.7529±0.0299 | 0.7375±0.0340 | 0.8367±0.0282 | 0.8439±0.0340 | 0.8463±0.0341 |

**Supplementary Table 4 | Performance of the XGBoost on IMP Diagnostic for grading task when using feature set obtained from different nuclear recognition methods.** The reported values are mean ± std of $mAP$. $C$ denotes, clinical features, $D_d$ denotes density-based features, $D_m$ denotes morphology-based features and $D_c$ denotes colocalisation features. $D$ refers to the combination of all types of features (excluding clinical) and $D$ is a subset of $\overline{D}$ after feature selection.

**Validation Set**

| | $\boldsymbol{D_d}$ | $\boldsymbol{D_m}$ | $\boldsymbol{D_c}$ | $\boldsymbol{D}$ | $\boldsymbol{\overline{D}}$ |
|---|---|---|---|---|---|
| **Baseline** | 0.7502±0.0263 | 0.8271±0.0261 | 0.8574±0.0164 | 0.8981±0.0183 | 0.8997±0.0175 |
| **Pathology AI** | 0.8234±0.0310 | 0.8472±0.0260 | 0.9209±0.0151 | 0.9302±0.0149 | 0.9349±0.0120 |
| **MDC Berlin \| IFP Bern** | 0.7770±0.0277 | 0.8114±0.0330 | 0.9188±0.0170 | 0.9357±0.0165 | 0.9385±0.0146 |
| **EPFL \| StarDist** | 0.8261±0.0278 | 0.8167±0.0253 | 0.9131±0.0146 | 0.9186±0.0137 | 0.9232±0.0134 |

**Testing Set**

| | $\boldsymbol{D_d}$ | $\boldsymbol{D_m}$ | $\boldsymbol{D_c}$ | $\boldsymbol{D}$ | $\boldsymbol{\overline{D}}$ |
|---|---|---|---|---|---|
| **Baseline** | 0.7456±0.0314 | 0.8189±0.0290 | 0.8547±0.0243 | 0.8921±0.0237 | 0.8956±0.0242 |
| **Pathology AI** | 0.8171±0.0328 | 0.8380±0.0302 | 0.9141±0.0235 | 0.9262±0.0227 | 0.9268±0.0219 |
| **MDC Berlin \| IFP Bern** | 0.7624±0.0295 | 0.8166±0.0322 | 0.9154±0.0189 | 0.9324±0.0191 | 0.9373±0.0179 |
| **EPFL \| StarDist** | 0.8175±0.0307 | 0.8136±0.0355 | 0.9058±0.0251 | 0.9142±0.0245 | 0.9185±0.0243 |

**Supplementary Table 5 | Performance of the XGBoost on IMP Diagnostic for grading task when using feature set obtained from different nuclear recognition methods.** The reported values are mean ± std of $QWK$ (Quadratic Weighted Kappa). $C$ denotes, clinical features, $D_d$ denotes density-based features, $D_m$ denotes morphology-based features and $D_c$ denotes colocalisation features. $D$ refers to the combination of all types of features (excluding clinical) and $D$ is a subset of $\overline{D}$ after feature selection.

**Validation Set**

| | $\boldsymbol{D_d}$ | $\boldsymbol{D_m}$ | $\boldsymbol{D_c}$ | $\boldsymbol{D}$ | $\boldsymbol{\overline{D}}$ |
|---|---|---|---|---|---|
| **Baseline** | 0.5892±0.0427 | 0.6539±0.0496 | 0.7074±0.0413 | 0.7600±0.0258 | 0.7696±0.0300 |
| **Pathology AI** | 0.6829±0.0308 | 0.7052±0.0445 | 0.8066±0.0349 | 0.8333±0.0299 | 0.8392±0.0243 |
| **MDC Berlin \| IFP Bern** | 0.6178±0.0447 | 0.6689±0.0538 | 0.8259±0.0361 | 0.8413±0.0335 | 0.8501±0.0330 |
| **EPFL \| StarDist** | 0.6974±0.0385 | 0.6786±0.0439 | 0.8082±0.0287 | 0.8193±0.0312 | 0.8248±0.0309 |

**Testing Set**

| | $\boldsymbol{D_d}$ | $\boldsymbol{D_m}$ | $\boldsymbol{D_c}$ | $\boldsymbol{D}$ | $\boldsymbol{\overline{D}}$ |
|---|---|---|---|---|---|
| **Baseline** | 0.5720±0.0580 | 0.6328±0.0551 | 0.6990±0.0406 | 0.7506±0.0551 | 0.7574±0.0492 |
| **Pathology AI** | 0.6696±0.0562 | 0.6904±0.0468 | 0.8051±0.0423 | 0.8302±0.0404 | 0.8354±0.0367 |
| **MDC Berlin \| IFP Bern** | 0.5842±0.0400 | 0.6685±0.0506 | 0.8212±0.0361 | 0.8436±0.0382 | 0.8463±0.0319 |
| **EPFL \| StarDist** | 0.6732±0.0467 | 0.6744±0.0480 | 0.7943±0.0383 | 0.8038±0.0477 | 0.8051±0.0477 |

# S4. Hyperparameter search for downstream tasks

We performed a random search over the XGBoost hyperparameters for each downstream clinical task to select the best model. This search space is defined as follows:

**Supplementary Table 6 | Hyperparameter space when performing Random Search.** We provide the name of the parameter, as used in the Python implementation (https://xgboost.readthedocs.io/en/stable/parameter.html), along with the range of values that we randomly sample from.

| Parameter Name | Value Ranges |
|---|---|
| num_boost_round | 8 to 256 |
| learning_rate | 0.001 to 0.1 |
| max_depth | 1 to 16 |
| subsample | One of [0.3, 0.4, 0.5, 0.6, 0.7, 0.8] |
| colsample_bytree | One of [0.3, 0.4, 0.5, 0.6, 0.7, 0.8] |
| colsample_bylevel | One of [0.3, 0.4, 0.5, 0.6, 0.7, 0.8] |
| colsample_bynode | One of [0.3, 0.4, 0.5, 0.6, 0.7, 0.8] |
| min_child_weight | 0.01 to 3.0 |
| reg_lambda | 0.1 to 2.0 |
| reg_alpha | 0.1 to 2.0 |
| booster | “booster” or “dart” |
| rate_drop | 0.1 to 0.7 |

We uniformly sampled the above search space to obtain 2048 parameter sets for subsequent analyses.

# S5. Complete feature description for downstream tasks

**Supplementary Table 7 | Complete list of features considered in the downstream pipelines.** Here, we give a description of the feature, along with the category in which it belongs.

| ID | Feature Names | Category |
|---|---|---|
| 0 | Average Connective's area | Morphology |
| 1 | Variation in Connective's area | Morphology |
| 2 | Average Connective's eccentricity | Morphology |
| 3 | Variation in Connective's eccentricity | Morphology |
| 4 | Average Connective's perimeter | Morphology |
| 5 | Variation in Connective's perimeter | Morphology |
| 6 | Average Connective's minor axis length | Morphology |
| 7 | Variation in Connective's minor axis length | Morphology |
| 8 | Average Connective's minor axis length | Morphology |
| 9 | Variation in Connective's minor axis length | Morphology |
| 10 | Average Connective's BAM | Morphology |
| 11 | Variation in Connective's BAM | Morphology |
| 12 | Average Eosinophil's area | Morphology |
| 13 | Variation in Eosinophil's area | Morphology |
| 14 | Average Eosinophil's eccentricity | Morphology |
| 15 | Variation in Eosinophil's eccentricity | Morphology |
| 16 | Average Eosinophil's perimeter | Morphology |
| 17 | Variation in Eosinophil's perimeter | Morphology |
| 18 | Average Eosinophil's minor axis length | Morphology |
| 19 | Variation in Eosinophil's minor axis length | Morphology |
| 20 | Average Eosinophil's minor axis length | Morphology |
| 21 | Variation in Eosinophil's minor axis length | Morphology |
| 22 | Average Eosinophil's BAM | Morphology |
| 23 | Variation in Eosinophil's BAM | Morphology |
| 24 | Average Epithelial's area | Morphology |
| 25 | Variation in Epithelial's area | Morphology |
| 26 | Average Epithelial's eccentricity | Morphology |
| 27 | Variation in Epithelial's eccentricity | Morphology |
| 28 | Average Epithelial's perimeter | Morphology |
| 29 | Variation in Epithelial's perimeter | Morphology |
| 30 | Average Epithelial's minor axis length | Morphology |
| 31 | Variation in Epithelial's minor axis length | Morphology |
| 32 | Average Epithelial's minor axis length | Morphology |
| 33 | Variation in Epithelial's minor axis length | Morphology |
| 34 | Average Epithelial's BAM | Morphology |
| 35 | Variation in Epithelial's BAM | Morphology |
| 36 | Average Lymphocyte's area | Morphology |
| 37 | Variation in Lymphocyte's area | Morphology |
| 38 | Average Lymphocyte's eccentricity | Morphology |
| 39 | Variation in Lymphocyte's eccentricity | Morphology |
| 40 | Average Lymphocyte's perimeter | Morphology |
| 41 | Variation in Lymphocyte's perimeter | Morphology |
| 42 | Average Lymphocyte's minor axis length | Morphology |
| 43 | Variation in Lymphocyte's minor axis length | Morphology |
| 44 | Average Lymphocyte's minor axis length | Morphology |
| 45 | Variation in Lymphocyte's minor axis length | Morphology |
| 46 | Average Lymphocyte's BAM | Morphology |
| 47 | Variation in Lymphocyte's BAM | Morphology |
| 48 | Average Neutrophil's area | Morphology |
| 49 | Variation in Neutrophil's area | Morphology |
| 50 | Average Neutrophil's eccentricity | Morphology |
| 51 | Variation in Neutrophil's eccentricity | Morphology |
| 52 | Average Neutrophil's perimeter | Morphology |
| 53 | Variation in Neutrophil's perimeter | Morphology |
| 54 | Average Neutrophil's minor axis length | Morphology |
| 55 | Variation in Neutrophil's minor axis length | Morphology |
| 56 | Average Neutrophil's minor axis length | Morphology |
| 57 | Variation in Neutrophil's minor axis length | Morphology |
| 58 | Average Neutrophil's BAM | Morphology |
| 59 | Variation in Neutrophil's BAM | Morphology |
| 60 | Average Plasma's area | Morphology |
| 61 | Variation in Plasma's area | Morphology |
| 62 | Average Plasma's eccentricity | Morphology |
| 63 | Variation in Plasma's eccentricity | Morphology |
| 64 | Average Plasma's perimeter | Morphology |
| 65 | Variation in Plasma's perimeter | Morphology |
| 66 | Average Plasma's minor axis length | Morphology |
| 67 | Variation in Plasma's minor axis length | Morphology |
| 68 | Average Plasma's minor axis length | Morphology |
| 69 | Variation in Plasma's minor axis length | Morphology |
| 70 | Average Plasma's BAM | Morphology |
| 71 | Variation in Plasma's BAM | Morphology |
| 72 | Average # Neutrophil within 200um radius of a Connective nucleus | Colocalisation |
| 73 | Variation in # Neutrophil within 200um radius of a Connective nucleus | Colocalisation |
| 74 | Average # Epithelial within 200um radius of a Connective nucleus | Colocalisation |
| 75 | Variation in # Epithelial within 200um radius of a Connective nucleus | Colocalisation |
| 76 | Average # Lymphocyte within 200um radius of a Connective nucleus | Colocalisation |
| 77 | Variation in # Lymphocyte within 200um radius of a Connective nucleus | Colocalisation |
| 78 | Average # Plasma within 200um radius of a Connective nucleus | Colocalisation |
| 79 | Variation in # Plasma within 200um radius of a Connective nucleus | Colocalisation |
| 80 | Average # Eosinophil within 200um radius of a Connective nucleus | Colocalisation |
| 81 | Variation in # Eosinophil within 200um radius of a Connective nucleus | Colocalisation |
| 82 | Average # Connective within 200um radius of a Connective nucleus | Colocalisation |
| 83 | Variation in # Connective within 200um radius of a Connective nucleus | Colocalisation |
| 84 | Average # Neutrophil within 200um radius of an Eosinophil nucleus | Colocalisation |
| 85 | Variation in # Neutrophil within 200um radius of an Eosinophil nucleus | Colocalisation |
| 86 | Average # Epithelial within 200um radius of an Eosinophil nucleus | Colocalisation |
| 87 | Variation in # Epithelial within 200um radius of an Eosinophil nucleus | Colocalisation |
| 88 | Average # Lymphocyte within 200um radius of an Eosinophil nucleus | Colocalisation |
| 89 | Variation in # Lymphocyte within 200um radius of an Eosinophil nucleus | Colocalisation |
| 90 | Average # Plasma within 200um radius of an Eosinophil nucleus | Colocalisation |
| 91 | Variation in # Plasma within 200um radius of an Eosinophil nucleus | Colocalisation |
| 92 | Average # Eosinophil within 200um radius of an Eosinophil nucleus | Colocalisation |
| 93 | Variation in # Eosinophil within 200um radius of an Eosinophil nucleus | Colocalisation |
| 94 | Average # Connective within 200um radius of an Eosinophil nucleus | Colocalisation |
| 95 | Variation in # Connective within 200um radius of an Eosinophil nucleus | Colocalisation |
| 96 | Average # Neutrophil within 200um radius of an Epithelial nucleus | Colocalisation |
| 97 | Variation in # Neutrophil within 200um radius of an Epithelial nucleus | Colocalisation |
| 98 | Average # Epithelial within 200um radius of an Epithelial nucleus | Colocalisation |
| 99 | Variation in # Epithelial within 200um radius of an Epithelial nucleus | Colocalisation |
| 100 | Average # Lymphocyte within 200um radius of an Epithelial nucleus | Colocalisation |
| 101 | Variation in # Lymphocyte within 200um radius of an Epithelial nucleus | Colocalisation |
| 102 | Average # Plasma within 200um radius of an Epithelial nucleus | Colocalisation |
| 103 | Variation in # Plasma within 200um radius of an Epithelial nucleus | Colocalisation |
| 104 | Average # Eosinophil within 200um radius of an Epithelial nucleus | Colocalisation |
| 105 | Variation in # Eosinophil within 200um radius of an Epithelial nucleus | Colocalisation |
| 106 | Average # Connective within 200um radius of an Epithelial nucleus | Colocalisation |
| 107 | Variation in # Connective within 200um radius of an Epithelial nucleus | Colocalisation |
| 108 | Average # Neutrophil within 200um radius of a Lymphocyte nucleus | Colocalisation |
| 109 | Variation in # Neutrophil within 200um radius of a Lymphocyte nucleus | Colocalisation |
| 110 | Average # Epithelial within 200um radius of a Lymphocyte nucleus | Colocalisation |
| 111 | Variation in # Epithelial within 200um radius of a Lymphocyte nucleus | Colocalisation |
| 112 | Average # Lymphocyte within 200um radius of a Lymphocyte nucleus | Colocalisation |
| 113 | Variation in # Lymphocyte within 200um radius of a Lymphocyte nucleus | Colocalisation |
| 114 | Average # Plasma within 200um radius of a Lymphocyte nucleus | Colocalisation |
| 115 | Variation in # Plasma within 200um radius of a Lymphocyte nucleus | Colocalisation |
| 116 | Average # Eosinophil within 200um radius of a Lymphocyte nucleus | Colocalisation |
| 117 | Variation in # Eosinophil within 200um radius of a Lymphocyte nucleus | Colocalisation |
| 118 | Average # Connective within 200um radius of a Lymphocyte nucleus | Colocalisation |
| 119 | Variation in # Connective within 200um radius of a Lymphocyte nucleus | Colocalisation |
| 120 | Average # Neutrophil within 200um radius of a Neutrophil nucleus | Colocalisation |
| 121 | Variation in # Neutrophil within 200um radius of a Neutrophil nucleus | Colocalisation |
| 122 | Average # Epithelial within 200um radius of a Neutrophil nucleus | Colocalisation |
| 123 | Variation in # Epithelial within 200um radius of a Neutrophil nucleus | Colocalisation |
| 124 | Average # Lymphocyte within 200um radius of a Neutrophil nucleus | Colocalisation |
| 125 | Variation in # Lymphocyte within 200um radius of a Neutrophil nucleus | Colocalisation |
| 126 | Average # Plasma within 200um radius of a Neutrophil nucleus | Colocalisation |
| 127 | Variation in # Plasma within 200um radius of a Neutrophil nucleus | Colocalisation |
| 128 | Average # Eosinophil within 200um radius of a Neutrophil nucleus | Colocalisation |
| 129 | Variation in # Eosinophil within 200um radius of a Neutrophil nucleus | Colocalisation |
| 130 | Average # Connective within 200um radius of a Neutrophil nucleus | Colocalisation |
| 131 | Variation in # Connective within 200um radius of a Neutrophil nucleus | Colocalisation |
| 132 | Average # Neutrophil within 200um radius of a Plasma nucleus | Colocalisation |
| 133 | Variation in # Neutrophil within 200um radius of a Plasma nucleus | Colocalisation |
| 134 | Average # Epithelial within 200um radius of a Plasma nucleus | Colocalisation |
| 135 | Variation in # Epithelial within 200um radius of a Plasma nucleus | Colocalisation |
| 136 | Average # Lymphocyte within 200um radius of a Plasma nucleus | Colocalisation |
| 137 | Variation in # Lymphocyte within 200um radius of a Plasma nucleus | Colocalisation |
| 138 | Average # Plasma within 200um radius of a Plasma nucleus | Colocalisation |
| 139 | Variation in # Plasma within 200um radius of a Plasma nucleus | Colocalisation |
| 140 | Average # Eosinophil within 200um radius of a Plasma nucleus | Colocalisation |
| 141 | Variation in # Eosinophil within 200um radius of a Plasma nucleus | Colocalisation |
| 142 | Average # Connective within 200um radius of a Plasma nucleus | Colocalisation |
| 143 | Variation in # Connective within 200um radius of a Plasma nucleus | Colocalisation |
| 144 | Average # Neutrophil within 400um radius of a Connective nucleus | Colocalisation |
| 145 | Variation in # Neutrophil within 400um radius of a Connective nucleus | Colocalisation |
| 146 | Average # Epithelial within 400um radius of a Connective nucleus | Colocalisation |
| 147 | Variation in # Epithelial within 400um radius of a Connective nucleus | Colocalisation |
| 148 | Average # Lymphocyte within 400um radius of a Connective nucleus | Colocalisation |
| 149 | Variation in # Lymphocyte within 400um radius of a Connective nucleus | Colocalisation |
| 150 | Average # Plasma within 400um radius of a Connective nucleus | Colocalisation |
| 151 | Variation in # Plasma within 400um radius of a Connective nucleus | Colocalisation |
| 152 | Average # Eosinophil within 400um radius of a Connective nucleus | Colocalisation |
| 153 | Variation in # Eosinophil within 400um radius of a Connective nucleus | Colocalisation |
| 154 | Average # Connective within 400um radius of a Connective nucleus | Colocalisation |
| 155 | Variation in # Connective within 400um radius of a Connective nucleus | Colocalisation |
| 156 | Average # Neutrophil within 400um radius of an Eosinophil nucleus | Colocalisation |
| 157 | Variation in # Neutrophil within 400um radius of an Eosinophil nucleus | Colocalisation |
| 158 | Average # Epithelial within 400um radius of an Eosinophil nucleus | Colocalisation |
| 159 | Variation in # Epithelial within 400um radius of an Eosinophil nucleus | Colocalisation |
| 160 | Average # Lymphocyte within 400um radius of an Eosinophil nucleus | Colocalisation |
| 161 | Variation in # Lymphocyte within 400um radius of an Eosinophil nucleus | Colocalisation |
| 162 | Average # Plasma within 400um radius of an Eosinophil nucleus | Colocalisation |
| 163 | Variation in # Plasma within 400um radius of an Eosinophil nucleus | Colocalisation |
| 164 | Average # Eosinophil within 400um radius of an Eosinophil nucleus | Colocalisation |
| 165 | Variation in # Eosinophil within 400um radius of an Eosinophil nucleus | Colocalisation |
| 166 | Average # Connective within 400um radius of an Eosinophil nucleus | Colocalisation |
| 167 | Variation in # Connective within 400um radius of an Eosinophil nucleus | Colocalisation |
| 168 | Average # Neutrophil within 400um radius of an Epithelial nucleus | Colocalisation |
| 169 | Variation in # Neutrophil within 400um radius of an Epithelial nucleus | Colocalisation |
| 170 | Average # Epithelial within 400um radius of an Epithelial nucleus | Colocalisation |
| 171 | Variation in # Epithelial within 400um radius of an Epithelial nucleus | Colocalisation |
| 172 | Average # Lymphocyte within 400um radius of an Epithelial nucleus | Colocalisation |
| 173 | Variation in # Lymphocyte within 400um radius of an Epithelial nucleus | Colocalisation |
| 174 | Average # Plasma within 400um radius of an Epithelial nucleus | Colocalisation |
| 175 | Variation in # Plasma within 400um radius of an Epithelial nucleus | Colocalisation |
| 176 | Average # Eosinophil within 400um radius of an Epithelial nucleus | Colocalisation |
| 177 | Variation in # Eosinophil within 400um radius of an Epithelial nucleus | Colocalisation |
| 178 | Average # Connective within 400um radius of an Epithelial nucleus | Colocalisation |
| 179 | Variation in # Connective within 400um radius of an Epithelial nucleus | Colocalisation |
| 180 | Average # Neutrophil within 400um radius of a Lymphocyte nucleus | Colocalisation |
| 181 | Variation in # Neutrophil within 400um radius of a Lymphocyte nucleus | Colocalisation |
| 182 | Average # Epithelial within 400um radius of a Lymphocyte nucleus | Colocalisation |
| 183 | Variation in # Epithelial within 400um radius of a Lymphocyte nucleus | Colocalisation |
| 184 | Average # Lymphocyte within 400um radius of a Lymphocyte nucleus | Colocalisation |
| 185 | Variation in # Lymphocyte within 400um radius of a Lymphocyte nucleus | Colocalisation |
| 186 | Average # Plasma within 400um radius of a Lymphocyte nucleus | Colocalisation |
| 187 | Variation in # Plasma within 400um radius of a Lymphocyte nucleus | Colocalisation |
| 188 | Average # Eosinophil within 400um radius of a Lymphocyte nucleus | Colocalisation |
| 189 | Variation in # Eosinophil within 400um radius of a Lymphocyte nucleus | Colocalisation |
| 190 | Average # Connective within 400um radius of a Lymphocyte nucleus | Colocalisation |
| 191 | Variation in # Connective within 400um radius of a Lymphocyte nucleus | Colocalisation |
| 192 | Average # Neutrophil within 400um radius of a Neutrophil nucleus | Colocalisation |
| 193 | Variation in # Neutrophil within 400um radius of a Neutrophil nucleus | Colocalisation |
| 194 | Average # Epithelial within 400um radius of a Neutrophil nucleus | Colocalisation |
| 195 | Variation in # Epithelial within 400um radius of a Neutrophil nucleus | Colocalisation |
| 196 | Average # Lymphocyte within 400um radius of a Neutrophil nucleus | Colocalisation |
| 197 | Variation in # Lymphocyte within 400um radius of a Neutrophil nucleus | Colocalisation |
| 198 | Average # Plasma within 400um radius of a Neutrophil nucleus | Colocalisation |
| 199 | Variation in # Plasma within 400um radius of a Neutrophil nucleus | Colocalisation |
| 200 | Average # Eosinophil within 400um radius of a Neutrophil nucleus | Colocalisation |
| 201 | Variation in # Eosinophil within 400um radius of a Neutrophil nucleus | Colocalisation |
| 202 | Average # Connective within 400um radius of a Neutrophil nucleus | Colocalisation |
| 203 | Variation in # Connective within 400um radius of a Neutrophil nucleus | Colocalisation |
| 204 | Average # Neutrophil within 400um radius of a Plasma nucleus | Colocalisation |
| 205 | Variation in # Neutrophil within 400um radius of a Plasma nucleus | Colocalisation |
| 206 | Average # Epithelial within 400um radius of a Plasma nucleus | Colocalisation |
| 207 | Variation in # Epithelial within 400um radius of a Plasma nucleus | Colocalisation |
| 208 | Average # Lymphocyte within 400um radius of a Plasma nucleus | Colocalisation |
| 209 | Variation in # Lymphocyte within 400um radius of a Plasma nucleus | Colocalisation |
| 210 | Average # Plasma within 400um radius of a Plasma nucleus | Colocalisation |
| 211 | Variation in # Plasma within 400um radius of a Plasma nucleus | Colocalisation |
| 212 | Average # Eosinophil within 400um radius of a Plasma nucleus | Colocalisation |
| 213 | Variation in # Eosinophil within 400um radius of a Plasma nucleus | Colocalisation |
| 214 | Average # Connective within 400um radius of a Plasma nucleus | Colocalisation |
| 215 | Variation in # Connective within 400um radius of a Plasma nucleus | Colocalisation |
| 216 | Connective cellular composition | Density |
| 217 | Eosinophil cellular composition | Density |
| 218 | Epithelial cellular composition | Density |
| 219 | Lymphocyte cellular composition | Density |
| 220 | Neutrophil cellular composition | Density |
| 221 | Plasma cellular composition | Density |

# Supplementary References

1. Schmidt, U., Weigert, M., Broaddus, C. & Myers, G. Cell detection with star-convex polygons. in *International Conference on Medical Image Computing and Computer-Assisted Intervention* 265-273 (Springer, 2018).
2. Ronneberger, O., Fischer, P. & Brox, T. U-net: Convolutional networks for biomedical image segmentation. in *International Conference on Medical image computing and computer-assisted intervention* 234-241 (Springer, 2015).
3. Tan, M. & Le, Q. Efficientnet: Rethinking model scaling for convolutional neural networks. in *International conference on machine learning* 6105-6114 (PMLR, 2019).
4. Graham, S.*, et al.* Hover-net: Simultaneous segmentation and classification of nuclei in multi-tissue histology images. *Medical Image Analysis* **58**, 101563 (2019).
5. Hu, J., Shen, L. & Sun, G. Squeeze-and-excitation networks. in *Proceedings of the IEEE conference on computer vision and pattern recognition* 7132-7141 (2018).
6. Srivastava, N., Hinton, G., Krizhevsky, A., Sutskever, I. & Salakhutdinov, R. Dropout: a simple way to prevent neural networks from overfitting. *The journal of machine learning research* **15**, 1929-1958 (2014).
7. Huang, G., Liu, Z., Van Der Maaten, L. & Weinberger, K.Q. Densely connected convolutional networks. in *Proceedings of the IEEE conference on computer vision and pattern recognition* 4700-4708 (2017).
8. Hou, Q., Zhou, D. & Feng, J. Coordinate attention for efficient mobile network design. in *Proceedings of the IEEE/CVF conference on computer vision and pattern recognition* 13713-13722 (2021).
9. Zhou, Z., Siddiquee, M.M.R., Tajbakhsh, N. & Liang, J. Unet++: Redesigning skip connections to exploit multiscale features in image segmentation. *IEEE transactions on medical imaging* **39**, 1856-1867 (2019).
10. Ghiasi, G.*, et al.* Simple copy-paste is a strong data augmentation method for instance segmentation. in *Proceedings of the IEEE/CVF Conference on Computer Vision and Pattern Recognition* 2918-2928 (2021).
11. DeVries, T. & Taylor, G.W. Improved regularization of convolutional neural networks with cutout. *arXiv preprint arXiv:1708.04552* (2017).
12. Yun, S.*, et al.* Cutmix: Regularization strategy to train strong classifiers with localizable features. in *Proceedings of the IEEE/CVF international conference on computer vision* 6023-6032 (2019).
13. Cai, Z. & Vasconcelos, N. Cascade r-cnn: Delving into high quality object detection. in *Proceedings of the IEEE conference on computer vision and pattern recognition* 6154-6162 (2018).
14. Xie, S., Girshick, R., Dollár, P., Tu, Z. & He, K. Aggregated residual transformations for deep neural networks. in *Proceedings of the IEEE conference on computer vision and pattern recognition* 1492-1500 (2017).
15. Wu, Y. & He, K. Group normalization. in *Proceedings of the European conference on computer vision (ECCV)* 3-19 (2018).
16. Misra, D. Mish: A self regularized non-monotonic neural activation function. *arXiv preprint arXiv:1908.08681* **4**, 10.48550 (2019).
17. Liu, Z.*, et al.* A convnet for the 2020s. in *Proceedings of the IEEE/CVF Conference on Computer Vision and Pattern Recognition* 11976-11986 (2022).
18. Yeung, M., Sala, E., Schönlieb, C.-B. & Rundo, L. Unified Focal loss: Generalising Dice and cross entropy-based losses to handle class imbalanced medical image segmentation. *Computerized Medical Imaging and Graphics* **95**, 102026 (2022).
19. Lin, T.-Y., Goyal, P., Girshick, R., He, K. & Dollár, P. Focal loss for dense object detection. in *Proceedings of the IEEE international conference on computer vision* 2980-2988 (2017).

20. Zhao, B., *et al.* Triple U-net: Hematoxylin-aware nuclei segmentation with progressive dense feature aggregation. *Medical Image Analysis* **65**, 101786 (2020).
21. Reinhard, E., Adhikhmin, M., Gooch, B. & Shirley, P. Color transfer between images. *IEEE Computer graphics and applications* **21**, 34-41 (2001).
22. Liu, Z., *et al.* Swin transformer: Hierarchical vision transformer using shifted windows. in *Proceedings of the IEEE/CVF International Conference on Computer Vision* 10012-10022 (2021).
23. Chen, K., *et al.* Hybrid task cascade for instance segmentation. in *Proceedings of the IEEE/CVF Conference on Computer Vision and Pattern Recognition* 4974-4983 (2019).
24. Macenko, M., *et al.* A method for normalizing histology slides for quantitative analysis. in *2009 IEEE international symposium on biomedical imaging: from nano to macro* 1107-1110 (IEEE, 2009).
25. Goodfellow, I., *et al.* Generative adversarial networks. *Communications of the ACM* **63**, 139-144 (2020).
26. Zhao, B., *et al.* RestainNet: a self-supervised digital re-stainer for stain normalization. *Computers and Electrical Engineering* **103**, 108304 (2022).
27. Duan, K., *et al.* Centernet: Keypoint triplets for object detection. in *Proceedings of the IEEE/CVF international conference on computer vision* 6569-6578 (2019).
28. He, K., Gkioxari, G., Dollár, P. & Girshick, R. Mask r-cnn. in *Proceedings of the IEEE international conference on computer vision* 2961-2969 (2017).
29. Sun, K., Xiao, B., Liu, D. & Wang, J. Deep high-resolution representation learning for human pose estimation. in *Proceedings of the IEEE/CVF conference on computer vision and pattern recognition* 5693-5703 (2019).
30. Chen, L.-C., Papandreou, G., Kokkinos, I., Murphy, K. & Yuille, A.L. Deeplab: Semantic image segmentation with deep convolutional nets, atrous convolution, and fully connected crfs. *IEEE transactions on pattern analysis and machine intelligence* **40**, 834-848 (2017).
31. Redmon, J., Divvala, S., Girshick, R. & Farhadi, A. You only look once: Unified, real-time object detection. in *Proceedings of the IEEE conference on computer vision and pattern recognition* 779-788 (2016).
32. Baheti, B., Innani, S., Gajre, S. & Talbar, S. Eff-unet: A novel architecture for semantic segmentation in unstructured environment. in *Proceedings of the IEEE/CVF Conference on Computer Vision and Pattern Recognition Workshops* 358-359 (2020).
33. Dawood, M., Branson, K., Rajpoot, N.M. & Minhas, F. Albrt: Cellular composition prediction in routine histology images. in *Proceedings of the IEEE/CVF International Conference on Computer Vision* 664-673 (2021).
34. Chollet, F. Xception: Deep learning with depthwise separable convolutions. in *Proceedings of the IEEE conference on computer vision and pattern recognition* 1251-1258 (2017).
35. Wang, C.-Y., *et al.* CSPNet: A new backbone that can enhance learning capability of CNN. in *Proceedings of the IEEE/CVF conference on computer vision and pattern recognition workshops* 390-391 (2020).